\documentclass{article}

\PassOptionsToPackage{numbers,sort}{natbib}
\usepackage[numbers]{natbib}
\usepackage[final]{neurips_2021}

\usepackage[utf8]{inputenc} 
\usepackage[T1]{fontenc}    
\usepackage[hidelinks]{hyperref}       
\usepackage{url}            
\usepackage{booktabs}       
\usepackage{amsfonts}       
\usepackage{nicefrac}       
\usepackage{microtype}      
\usepackage{xcolor}         

\title{Federated Hyperparameter Tuning: Challenges, Baselines, and Connections to Weight-Sharing}

\author{%
  Mikhail Khodak, Renbo Tu, Tian Li\\
  Carnegie Mellon University\\
  \texttt{\{khodak,renbo,tianli\}@cmu.edu} \\
  \And Liam Li\\
  Hewlett Packard Enterprise\\
  \texttt{me@liamcli.com} \\
  \And Maria-Florina Balcan, Virginia Smith\\
  Carnegie Mellon University\\
  \texttt{ninamf@cs.cmu.edu,smithv@cmu.edu}
  \And Ameet Talwalkar \\
  Carnegie Mellon University \& Hewlett Packard Enterprise\\
  \texttt{talwalkar@cmu.edu}
}

\usepackage{graphicx}
\usepackage{subfigure}

\usepackage{amsmath,amsthm}
\usepackage[ruled,vlined]{algorithm2e}
\usepackage[utf8]{inputenc}
\usepackage[hang,flushmargin]{footmisc}
\usepackage{threeparttable,multirow}
\usepackage{tikz}
\usetikzlibrary{tikzmark,calc}

\newtheorem{Thm}{Theorem}[section]

\DeclareMathOperator*{\argmin}{arg\,min}

\newcommand{\Unif}{\operatorname{Unif}}

\newcommand{\A}{\mathcal A}
\newcommand{\B}{\mathcal B}
\newcommand{\Config}{\mathcal C}
\newcommand{\D}{\mathcal D}
\newcommand{\E}{\mathbb E}

\newcommand{\R}{\mathbb R}
\newcommand{\W}{\mathcal W}

\newcommand{\Z}{\mathbb Z}

\usepackage{scalerel,mathtools,color}
\let\svsqrt\sqrt
\newsavebox\Nsqrt
\def\sr#1{\ThisStyle{%
		\savebox\Nsqrt{\scalebox{.5}[1]{$\SavedStyle\svsqrt{\phantom{\cramped{#1#1}}}$}}%
		\ooalign{\usebox{\Nsqrt}\cr\kern.2pt\usebox{\Nsqrt}\cr\hfil$\SavedStyle\cramped{#1}$}}}

\usepackage{enumitem}
\def\*#1{\mathbf{#1}}

\newcommand{\SGD}{\operatorname{\texttt{Loc}}}
\newcommand{\Alg}{\operatorname{\texttt{Alg}}}
\newcommand{\Agg}{\operatorname{\texttt{Agg}}}

\newcommand{\FedEx}{\texttt{FedEx}\xspace}
\newcommand{\fedex}{\texttt{FedEx}\xspace}
\newcommand{\FedProx}{\texttt{FedProx}\xspace}

\newcommand{\FedAvg}{\texttt{FedAvg}\xspace}
\newcommand{\SCAFFOLD}{\texttt{SCAFFOLD}\xspace}

\newcommand{\Reptile}{\texttt{Reptile}\xspace}
\newcommand{\MAML}{\texttt{MAML}\xspace}
\newcommand{\1}{\mathbf{1}}
\newcommand{\diam}{\operatorname{diam}}

\begin{document}

\maketitle

\begin{abstract}
	Tuning hyperparameters is a crucial but arduous part of the machine learning  pipeline.
	Hyperparameter optimization is even more challenging in federated learning, 
	where models are learned over a distributed network of heterogeneous devices; here, the need to keep data on device and perform local training makes it difficult to efficiently train and evaluate configurations.  
	In this work, we investigate the problem of federated hyperparameter tuning. 
	We first identify key challenges and show how standard approaches may be adapted to form baselines for the federated setting.
	Then, by making a novel connection to the neural architecture search 
	technique of {\em weight-sharing}, we introduce a new method, \FedEx, to accelerate federated hyperparameter tuning
	that is applicable to widely-used federated optimization methods such as \FedAvg and recent variants.
	Theoretically, we show that a \FedEx variant correctly tunes the on-device learning rate in the setting of online convex optimization across devices.
	Empirically, we show that \fedex can outperform natural baselines for federated hyperparameter tuning by several percentage points on the Shakespeare, FEMNIST, and CIFAR-10 benchmarks---obtaining higher accuracy using the same training budget.
\end{abstract}


\section{Introduction}\label{sec:intro}
\vspace{-1mm}

Federated learning (FL) is a popular distributed computational setting where training is performed locally or privately~\citep{mcmahan:17,li:20a} and where hyperparameter tuning has been identified as a critical problem~\citep{kairouz:19}. Although general hyperparameter optimization has been the subject of intense study~\citep{hutter:11,bergstra:12,li:18}, several unique aspects of the federated setting make tuning hyperparameters especially challenging. However, to the best of our knowledge there has been no dedicated study on the specific challenges and solutions in federated hyperparameter tuning. 
In this work, we first formalize the problem of hyperparameter optimization in FL, introducing the following three key challenges:
\begin{enumerate}[leftmargin=*,topsep=-1pt,noitemsep]\setlength\itemsep{2pt}
	\item {\bf Federated validation data:} In federated networks, as the validation data is split across devices, the entire dataset is not available at any one time;
	instead a central server is given access to some number of devices at each communication round, for one or at most a few runs of local training and validation.
	Thus, because the standard measure of complexity in FL is the number of communication rounds, computing validation metrics exactly dramatically increases the cost.
	\item {\bf Extreme resource limitations:} 
	FL applications often involve training using devices with very limited computational and communication capabilities.
	Furthermore, many require the use of privacy techniques such as differential privacy that limit the number times user data can be accessed.
	Thus we cannot depend on being able to run many different configurations to completion.
	\item {\bf Evaluating personalization:} Finally, even with non-federated data, applying common hyperparameter optimization methods to standard personalized FL approaches (such as finetuning) can be costly because evaluation may require performing many additional training steps locally.
\end{enumerate}
With these challenges\footnote{A further challenge we do {\em not} address is that of the time-dependency of federated evaluation, c.f. \cite{eichner:19}.} in mind, we propose reasonable baselines for federated hyperparameter tuning by showing how to adapt standard non-federated algorithms.
We further study the challenge of noisy validation signal due to federation, and show that simple state-estimation-based fixes do not help.

Our formalization and analysis of this problem leads us to develop \FedEx, a method that exploits a novel connection between
hyperparameter tuning in FL and the weight-sharing technique widely used in neural architecture search (NAS) \citep{pham:18,liu:19,cai:19}. In particular, we observe that weight-sharing is a natural way of addressing the three challenges above for federated hyperparameter tuning, as it incorporates noisy validation signal, simultaneously tunes and trains the model, and evaluates personalization as part of training rather than as a costly separate step. 
Although standard weight-sharing only handles architectural hyperparameters such as the choice of layer or activation, and not critical settings such as those of local stochastic gradient descent (SGD), we develop a formulation that allows us to tune most of these as well via the relationship between local-training and fine-tuning-based personalization.
This make \FedEx a general hyperparameter tuning algorithm applicable to many local training-based FL methods, e.g. \FedAvg \citep{mcmahan:17}, \FedProx \citep{li:20b}, and \SCAFFOLD \citep{karimireddy:19}.

In Section~\ref{sec:theory}, we next conduct a theoretical study of \FedEx  in a simple setting: tuning the client step-size. Using the ARUBA framework for analyzing meta-learning \citep{khodak:19b}, we show that a variant of \FedEx correctly tunes the on-device step-size to minimize client-averaged regret by adapting to the intrinsic similarity between client data.
We improve the convergence rate compared to some past meta-learning theory~\citep{khodak:19b,li:20c} while not depending on knowing the (usually unknown) task-similarity.

Finally, in Section~\ref{sec:empirical}, we instantiate our baselines and \FedEx to tune hyperparameters of \FedAvg, \FedProx, and \Reptile, evaluating 
 on three standard FL benchmarks:
Shakespeare, FEMNIST, and CIFAR-10 \citep{mcmahan:17,caldas:18}.
While our baselines already obtain performance similar to past hand-tuning, \FedEx further surpasses them in most settings examined, including by 2-3\% on Shakespeare.

\begin{figure}[t!]
	\centering
	\includegraphics[width=0.59\textwidth]{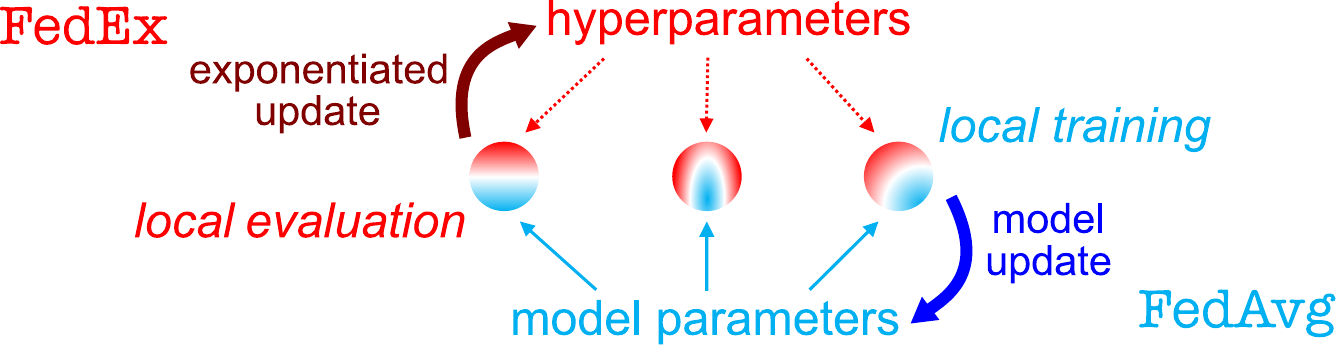}
	\vspace{-2mm}
	\caption{
		\FedEx can be applied to any local training-based FL method, e.g. \FedAvg, by interleaving standard updates to model weights (computed by aggregating results of local training) with exponentiated gradient updates to hyperparameters (computed by aggregating results of local validation).}
	\label{fig:overview}
\end{figure}

\vspace{-1mm}
\paragraph{Related Work}
To the best of our knowledge, we are the first to systematically analyze the formulation and challenges of hyperparameter optimization in the federated setting. 
Several papers have explored limited aspects of hyperparameter tuning in FL~\citep{mostafa:19,koskela:18,dai:20}, focusing on a small number of hyperparameters (e.g. the step-size and sometimes one or two more) in less general settings (studying small-scale problems or assuming server-side validation data).
In contrast our methods are able to tune a wide range of hyperparameters in realistic federated networks.
Some papers also discussed the challenges of finding good configurations while studying other aspects of federated training \citep{reddi:21}.
We argue that it is critical to properly address the challenges of federated hyperparameter optimization in practical settings, as we discuss in detail in Section~\ref{sec:federated}.

Methodologically, our approach draws on the fact that local training-based methods such as \FedAvg can be viewed as optimizing a surrogate objective for personalization~\citep{khodak:19b}, and more broadly leverages the similarity of the personalized FL setup and initialization-based meta-learning \citep{chen:18,li:20c,jiang:19,fallah:20b}.
While \FedEx's formulation and guarantees use this relationship, the method itself is general-purpose and applicable to federated training of a single global model.
Many recent papers address FL personalization more directly~\citep{mansour2020three,yu2020salvaging,ghosh2020efficient,smith:17,li2020ditto}.
This connection and our use of NAS techniques also makes research connecting NAS and meta-learning relevant \citep{lian:20,elsken:19}, but unlike these methods we focus on tuning {\em non}-architectural parameters.
In fact, we believe our work is the first to apply weight-sharing to regular hyperparameter search.
Furthermore, meta-learning does not have the data-access and computational restrictions of FL, where such methods using the DARTS mixture relaxation~\citep{liu:19} are less practical.
Instead, \FedEx employs the lower-overhead stochastic relaxation~\citep{li:19a,dong:19}, and its exponentiated update is similar to the recently proposed GAEA approach for NAS \citep{li:21}.
Running NAS itself in federated settings has also been studied \citep{garg:20,he:20,xu:20b};
while our focus is on non-architectural hyperparameters, in-principle our algorithms can also be used for federated NAS.

Theoretically, our work makes use of the average regret-upper-bound analysis (ARUBA) framework~\citep{khodak:19b} to derive guarantees for learning the initialization, i.e. the global model, while simultaneously tuning the step-size of the local algorithm.
The step-size of gradient-based algorithms has also been tuned on its own in the settings of data-driven algorithm design \citep{gupta2017pac} and of statistical learning-to-learn~\citep{wang2021guarantees}.

\section{Federated Hyperparameter Optimization}\label{sec:federated}
\vspace{-1mm}

In this section we formalize the problem of hyperparameter optimization for FL and discuss the connection of its personalized variant to meta-learning.
We also review \FedAvg~\citep{mcmahan:17}, a common federated optimization method, and present a reasonable baseline approach for tuning its hyperparameters.

\vspace{-1mm}
\paragraph{Global and Personalized FL}
In FL we are concerned with optimizing over a network of heterogeneous clients $i=1,\dots,n$, each with training, validation, and testing sets $T_i$, $V_i$, and $E_i$, respectively.
We use $L_S(\*w)$ to denote the average loss over a dataset $S$ of some $\*w$-parameterized ML model, for $\*w\in\R^d$ some real vector.
For hyperparameter optimization, we assume a class of algorithms $\Alg_a$ hyperparameterized by $a\in\A$ that use {\em federated access} to training sets $T_i$ to output some element of $\R^d$.
Here by ``federated access" we mean that each iteration corresponds to a {\em communication round} at which $\Alg_a$ has access to a batch of $B$ clients\footnote{For simplicity the number of clients per round is fixed, but all methods can be easily generalized to varying $B$.} that can do local training and validation.

Specifically, we assume $\Alg_a$ can be described by two subroutines with hyperparameters encoded by $b\in\B$ and $c\in\Config$, so that $a=(b,c)$ and $\A=\B\times\Config$. Here
$c$ encodes settings of a local training algorithm $\SGD_c$ that take a training set $S$ and initialization $\*w\in\R^d$ as input and outputs a model $\SGD_c(S,\*w)\in\R^d$, while $b$ sets those of an aggregation $\Agg_b$ that takes the initialization $\*w$ and outputs of $\SGD_c$ as input and returns a model parameter.
For example, in standard \FedAvg, $\SGD_c$ is $T$ steps of gradient descent with step-size $\eta$ and $\Agg_b$ takes a weighted average of the outputs of $\SGD_c$ across clients;
here $c=(\eta,T)$ and $b=()$.
As detailed in the appendix, many FL methods can be decomposed this way, including well-known ones such as \FedAvg~\cite{mcmahan:17}, \FedProx~\citep{li:20b}, \SCAFFOLD~\cite{karimireddy:19}, and \Reptile~\cite{nichol:18} as well as more recent methods \cite{li2020ditto,al-shedivat:21,acar2021federated}.
Our analysis and our proposed \FedEx algorithm will thus apply to all of them, up to an assumption detailed next.

Starting from this decomposition, the global hyperparameter optimization problem can be written as\vspace{-1mm}
\begin{equation}\label{eq:global}
\min_{a\in\A}\quad\sum_{i=1}^n|V_i|L_{V_i}(\Alg_a(\{T_j\}_{j=1}^n))
\end{equation}
In many cases we are also interested in obtaining a device-specific local model, where we take a model trained on all clients and finetune it on each individual client before evaluating.
A key assumption we make is that the finetuning algorithm will be the same as the local training algorithm $\SGD_c$ used by $\Alg_a$.
This assumption can be justified by recent work in meta-learning that shows that algorithms that aggregate the outputs of local SGD can be viewed as optimizing for personalization using local SGD \citep{khodak:19b}.
Then, in the personalized setting, the tuning objective becomes\vspace{-1mm}
\begin{equation}\label{eq:refine}
\min_{a=(b,c)\in\A}\quad\sum_{i=1}^n|V_i|L_{V_i}(\SGD_c(T_i,\Alg_a(\{T_j\}_{j=1}^n))
\end{equation}
Our approach will focus on the setting where the hyperparameters $c$ of local training make up a significant portion of all hyperparameters $a=(b,c)$;
by considering the personalization objective we will be able to treat such hyperparameters as architectural and thus apply weight-sharing.

\vspace{-1mm}
\paragraph{Tuning FL Methods: Challenges and Baselines}

\begin{figure*}[!t]
	\begin{minipage}{0.58\linewidth}
		\begin{algorithm}[H]
			\DontPrintSemicolon
			\KwIn{distribution $\D$ over hyperparameters $\A$, elimination rate $\eta\in\mathbb N$, elimination rounds $\tau_0=0,\tau_1,\dots,\tau_R$}
			sample set of $\eta^R$ hyperparameters $H\sim\D^{[\eta^R]}$\\
			initialize a model $\*w_a\in\R^d$ for each $a\in H$\\
			\For{elimination round $r\in[R]$}{
				\For{setting $a=(b,c)\in H$}{
					\For{comm. round $t=\tau_{r-1}+1,\dots,\tau_r$}{
						\For{client $i=1,\dots,B$}{
							send $\*w_a,c$ to client\\
							$\*w_i\gets\SGD_c(T_{ti},\*w_a)$\\
							send $\*w_i,L_{V_{ti}}(\*w_i)$ to server
						}
						$\*w_a\gets\Agg_b(\*w_a,\{\*w_i\}_{i=1}^B)$\\
						$s_a\gets\sum_{i=1}^B|V_{ti}|L_{V_{ti}}(\*w_i)/\sum_{i=1}^B|V_{ti}|$
					}
				}
				$H\gets\{a\in H:s_a\le\frac1\eta\textrm{-quantile}(\{s_a:a\in H\})\}$
			}
			\KwOut{remaining $a\in H$ and associated model $\*w_a$}
			\caption{\label{alg:baseline}
				Successive halving algorithm (SHA) applied to personalized FL.
				For the non-personalized objective \eqref{eq:global}, replace $L_{V_{ti}}(\*w_i)$ by $L_{V_{ti}}(\*w_a)$.
				For random search (RS) with $N$ samples, set $\eta=N$ and $R=1$.
			}
		\end{algorithm}
	\end{minipage}
	\begin{minipage}{0.42\linewidth}
		\centering
		\includegraphics[width=0.666\linewidth]{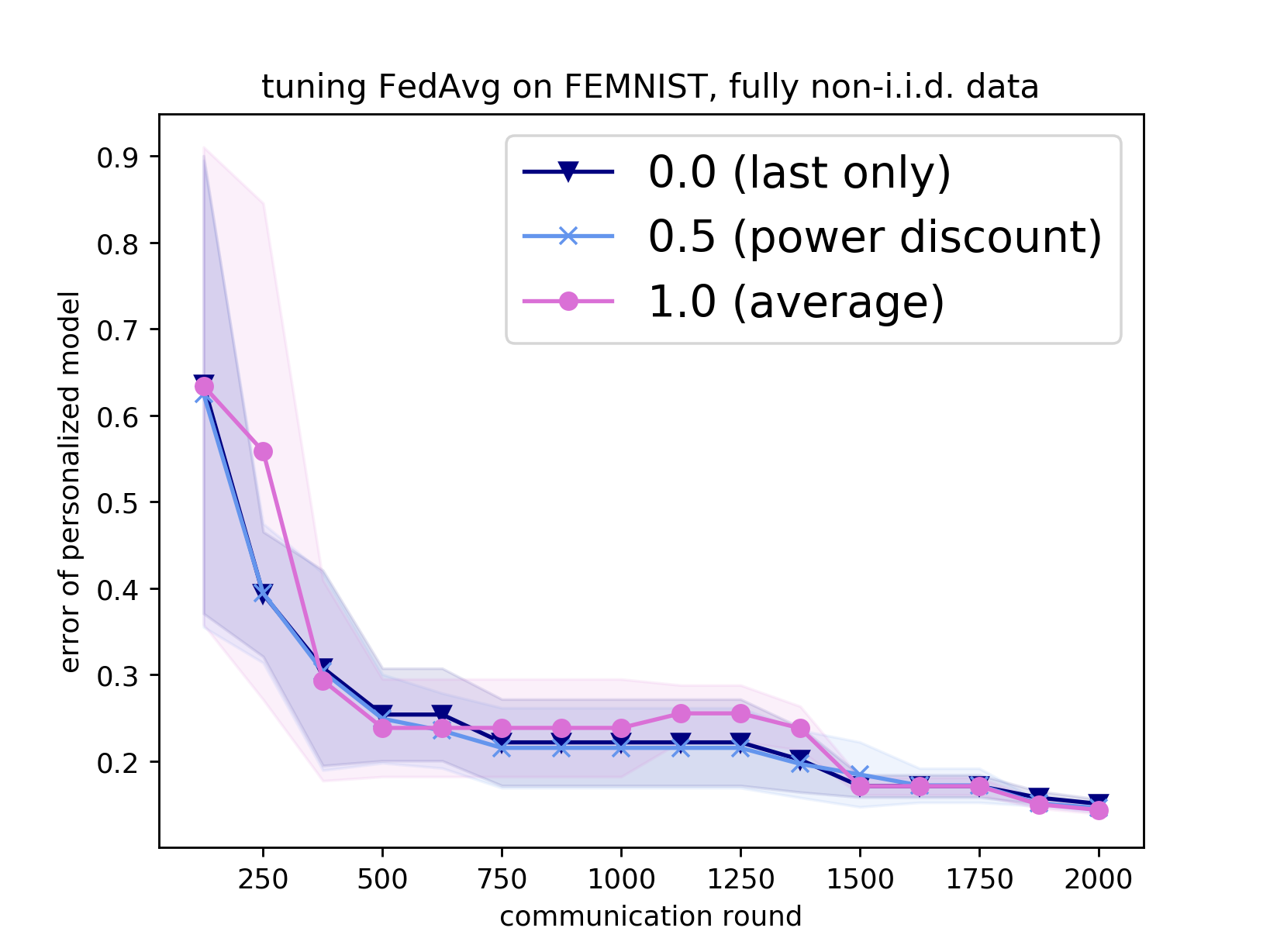}
		\hfill
		\includegraphics[width=0.666\linewidth]{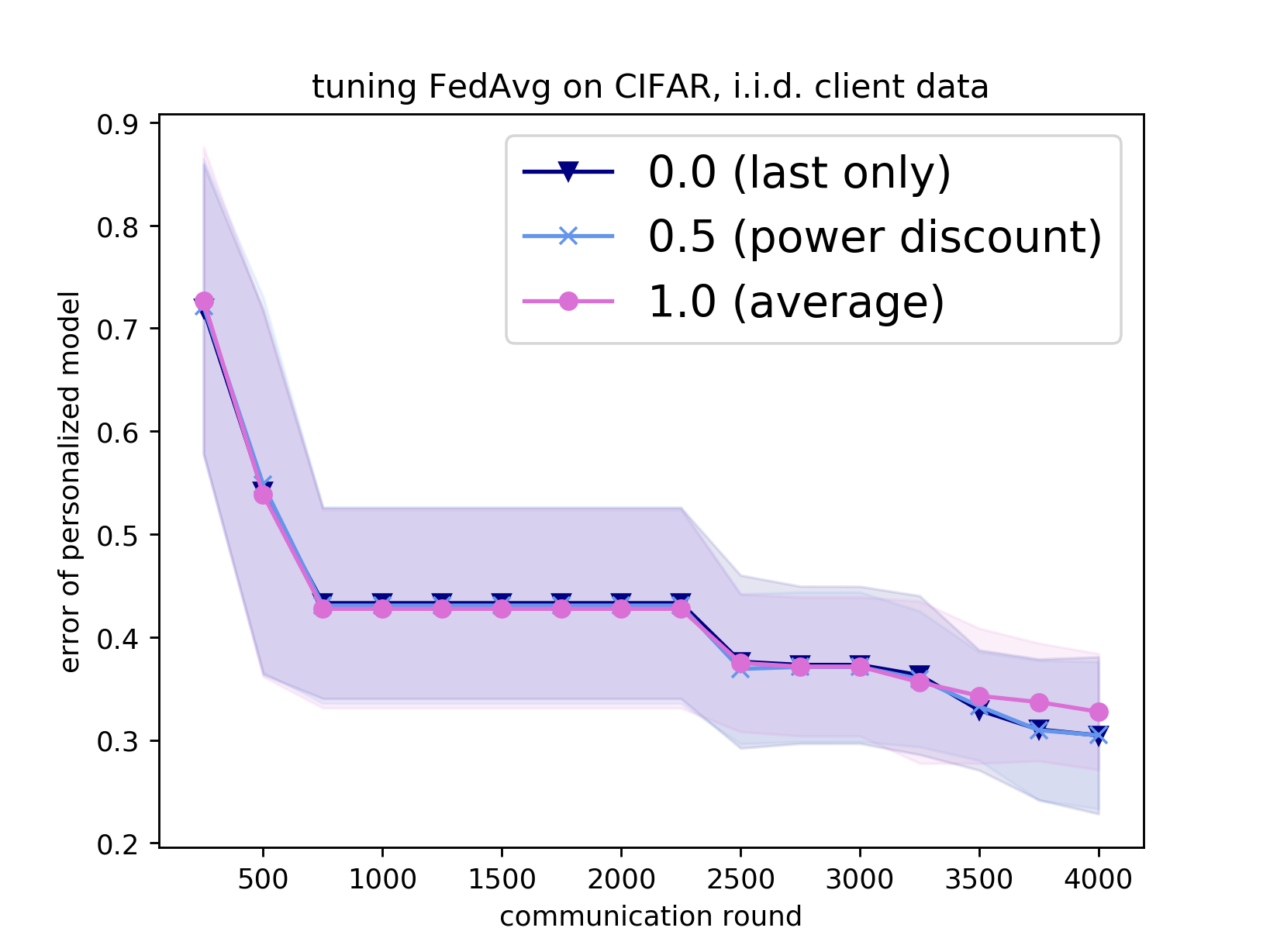}
		\caption{\label{fig:discount}
			Tuning FL with SHA but making elimination decisions based on validation estimates using different discount factors.
			On both FEMNIST (top) and CIFAR (bottom) using more of the validation data does not improve upon just using the most recent round's validation error.
		}
	\end{minipage}
\end{figure*}

In the non-federated setting, the objective~\eqref{eq:global} is amenable to regular hyperparameter optimization methods;
for example, a random search approach would repeatedly sample a setting $a$ from some distribution over $\A$, run $\Alg_a$ to completion, and evaluate the objective, saving the best setting and output \citep{bergstra:12}.
With a reasonable distribution and enough samples this is guaranteed to converge and can be accelerated using early stopping methods~\citep{li:18}, in which $\Alg_a$ is not always run to completion if the desired objective is poor at intermediate stages, or by adapting the sampling distribution using the results of previous objective evaluations \citep{snoek:12}.
As mentioned in the introduction, applying such methods to FL is inherently challenging due to
\begin{enumerate}[leftmargin=*,topsep=-1pt,noitemsep]\setlength\itemsep{2pt}
	\item {\bf Federated validation data:} Separating data across devices means we cannot immediately get a good estimate of the model's validation performance, as we only have access to a possibly small batch of devices at a time.
	This means that decisions such as which models to flag for early stopping 
	 will be noisy and may not fully incorporate all the available validation signal.
	\item {\bf Extreme resource limitations:} 
	As FL algorithms can take a very long time to run in-practice due to the weakness and spotty availability of devices, we often cannot afford to conduct many training runs to evaluate different configurations.
	This issue is made more salient in cases where we use privacy techniques that only allow a limited number of accesses to the data of any individual user.
	\item {\bf Evaluating personalization:} 
	While personalization is important in FL due to client heterogeneity, checking the performance of the current model on the personalization objective~\eqref{eq:refine} is computationally intensive because computing may require running local training multiple times.
	In particular, while regular validation losses require computing one forward pass per data point, personalized losses require several forward-backward passes, making it many times more expensive if this loss is needed to make a tuning decision such as eliminating a configuration from consideration.
\end{enumerate}

Despite these challenges, we can still devise sensible baselines for tuning hyperparameters in FL, most straightforward of which is to use a regular hyperparameter method but use validation data from a single round as a noisy surrogate for the full validation objective.
Specifically, one can use random search (RS)---repeatedly evaluate random configurations---and a simple generalization called successive halving (SHA), in which we sample a set of configurations and partially run all of them for some number of communication rounds before eliminating all but the best $\frac1\eta$ fraction, repeating until only one configuration remains.
Note both are equivalent to a ``bracket'' in Hyperband~\citep{li:18} and their adaptation to FL is detailed in Algorithm~\ref{alg:baseline}.

As shown in Section~\ref{sec:empirical}, SHA performs reasonably well on the benchmarks we consider.
However, by using validation data from one round it may make noisy elimination decisions, early-stopping potentially good configurations because of a difficult set of clients on a particular round.
Here the problem is one of insufficient utilization of the validation data to estimate model performance.
A reasonable approach to use more is to try some type of state-estimation:
using the performance from previous rounds to improve the noisy measurement of the current one.
For example, instead of using only the most recent round for elimination decisions we can use a weighted sum of the performances at all past rounds.
To investigate this, we study a power decay weighting, where a round is discounted by some constant factor for each time step it is in the past.
We consider factors 0.0 (taking the most recent performance only, as before), 0.5, and 1.0 (taking the average). 
However, in Figure~\ref{fig:discount} we show that incorporating more validation data this way than is used by Algorithm~\ref{alg:baseline} by default does not significantly affect results.

Thus we may need a better algorithm to use more of the validation signal, most of which is discarded by using the most recent round's performance.
We next propose \FedEx, a new method that does so by using validation on each round to update a client hyperparameters distribution used to sample configurations to send to devices.
Thus it alleviates issue~(1) above by updating at each step, not waiting for an elimination round as in RS or SHA.
By simultaneously training the model and tuning (client) hyperparameters, it also moves towards a fully single-shot procedure in which we only train once (we must still run multiple times due to server hyperparameters), which would solve issue~(2).
Finally, \FedEx addresses issue~(3) by using local training to both update the model and to estimate personalized validation loss, thus not spending extra computation on this more expensive objective.

\section{Weight-Sharing for Federated Learning}\label{sec:ws}
\vspace{-1mm}

We now present \FedEx, a way to tune local FL hyperparameters.
This section contains the general algorithm and its connection to weight-sharing; we instantiate it on several FL methods in Section~\ref{sec:empirical}.

\vspace{-1mm}
\paragraph{Weight-Sharing for Architecture Search}
We first review the weight-sharing approach in NAS, which for a set $\Config$ of network configurations is often posed as the bilevel optimization
\begin{equation}\label{eq:bilevel}
\min_{c\in\Config}~L_\textrm{valid}(\*w,c)\quad\textrm{s.t.}\quad\*w\in\argmin_{\*u\in\R^d}~L_\textrm{train}(\*u,c)
\end{equation}
where $L_\textrm{train},L_\textrm{valid}$ evaluate a single configuration with the given weights.
If, as in NAS, all hyperparameters are architectural, then they are effectively themselves trainable model parameters \citep{li:21}, so we could instead consider solving the following ``single-level" empirical risk minimization (ERM):
\begin{equation}\label{eq:erm}
\min_{c\in\Config,\*w\in\R^d}~L(\*w,c)\quad=\quad\min_{c\in\Config,\*w\in\R^d}~L_\textrm{train}(\*w,c)+L_\textrm{valid}(\*w,c)
\end{equation}
Solving this instead of the bilevel problem \eqref{eq:bilevel} has been proposed in several recent papers~\citep{li:19b,li:21}.

Early approaches to solving either formulation of NAS were costly due to the need for full or partial training of many architectures in a very large search space.
The weight-sharing paradigm \citep{pham:18} reduces the problem to that of training a single architecture, a ``supernet" containing all architectures in the search space $\Config$.
A straightforward way of constructing a supernet is via a ``stochastic relaxation" where the loss is an expectation w.r.t. sampling $c$ from some distribution over $\Config$ \citep{dong:19}.
Then the shared weights can be updated using SGD by first sampling an architecture $c$ and using an unbiased estimate of $\nabla_{\*w}L(\*w,c)$ to update $\*w$.
The distribution over $\Config$ may itself be adapted or stay fixed.
We focus on the former case, adapting some $\theta$-parameterized distribution $\D_\theta$; this yields the stochastic relaxation objective
\begin{equation}\label{eq:relaxed}
\min_{\theta\in\Theta,\*w\in\R^d}~\E_{c\sim\D_\theta}L(\*w,c)
\end{equation}
Since architectural hyperparameters are often discrete decisions, e.g. a choice of which of a fixed number of operations to use, a natural choice of $\D_\theta$ is as a product of categorical distributions over simplices.
In this case, any discretization of an optimum $\theta$ of the relaxed objective \eqref{eq:relaxed} whose support is in the support of $\theta$ will be an optimum of the original objective \eqref{eq:erm}.
A natural update scheme here is exponentiated gradient \citep{kivinen:97}, where each successive $\theta$ is proportional to $\theta\odot\exp(-\eta\tilde\nabla)$, $\eta$ is a step-size, and $\tilde\nabla$ an unbiased estimate of $\nabla_\theta\E_{c\sim\D_\theta}L(\*w,c)$ that can be computed using the re-parameterization trick \citep{rubinstein:93}.
By alternating this exponentiated update with the standard SGD update to $\*w$ discussed earlier we obtain a simple block-stochastic minimization scheme that is guaranteed to converge, under certain conditions, to the ERM objective, and also performs well in practice~\citep{li:21}.

\vspace{-1mm}
\paragraph{The \FedEx Method}
To obtain \FedEx from weight-sharing we restrict to the case of tuning only the hyperparameters $c$ of local training $\SGD_c$.\footnote{We will use some wrapper algorithm to tune the hyperparameters $b$ of $\Agg_b$.}
Our goal then is just to find the best initialization $\*w\in\R^d$ and local hyperparameters $c\in\Config$, i.e. we replace the personalized objective \eqref{eq:refine} by\vspace{-1mm}
\begin{equation}\label{eq:objective}
\min_{c\in\Config,\*w\in\R^d}\quad\sum_{i=1}^n|V_i|L_{V_i}(\SGD_c(T_i,\*w))\vspace{-1mm}
\end{equation}
Note $\Alg_a$ outputs an element of $\R^d$, so this new objective is upper-bounded by the original~\eqref{eq:refine}, i.e. any solution will be at least as good for the original objective.
Note also that for fixed $c$ this is equivalent to the classic train-validation split objective for meta-learning with $\SGD_c$ as the base-learner.
More importantly for us, it is also in the form of the r.h.s. of the weight-sharing objective~\eqref{eq:erm}, i.e. it is a single-level function of $\*w$ and $c$.
We thus apply a NAS-like stochastic relaxation:\vspace{-1mm}
\begin{equation}\label{eq:stochastic}
\min_{\theta\in\Theta,\*w\in\R^d}\quad\sum_{i=1}^n|V_i|\E_{c\in\D_\theta}L_{V_i}(\SGD_c(T_i,\*w))
\end{equation}
In NAS we would now set the distribution to be a product of categorical distributions over different architectures, thus making $\theta$ an element of a product of simplices and making the optimum of the original objective \eqref{eq:objective} equivalent to the optimum of the relaxed objective \eqref{eq:stochastic} as an extreme point of the simplex.
Unlike in NAS, FL hyperparameters such as the learning rate are not extreme points of a simplex and so it is less clear what parameterized distribution $\D_\theta$ to use.
Nevertheless, we find that crudely imposing a categorical distribution over $k>1$ random samples from some distribution (e.g. uniform) over $\Config$ and updating $\theta$ using exponentiated gradient over the resulting $k$-simplex works well.
We alternate this with updating $\*w\in\R^d$, which in a NAS algorithm involves an SGD update using an unbiased estimate of the gradient at the current $\*w$ and $\theta$.

\begin{figure}[!t]
\begin{minipage}{0.55\linewidth}
\begin{algorithm}[H]
	\DontPrintSemicolon
	\KwIn{configurations $c_1,\dots,c_k\in\Config$, setting $b$ for $\Agg_b$, schemes for setting step-size $\eta_t$ and baseline $\lambda_t$, total number of steps $\tau\ge1$}
	initialize $\theta_1=\1_k/k$ and shared weights $\*w_1\in\R^d$\\
	\For{comm. round $t=1,\dots,\tau$}{
		\For{client $i=1,\dots,B$}{
			send $\*w_t,\theta_t$ to client\\
			sample $c_{ti}\sim\D_{\theta_t}$\\
			$\*w_{ti}\gets\SGD_{c_{ti}}(T_{ti},\*w_t)$\\
			send $\*w_{ti},c_{ti},L_{V_{ti}}(\*w_{ti})$ to server
		}
		$\*w_{t+1}\gets\Agg_b(\*w,\{\*w_{ti}\}_{i=1}^B)$\\
 $\tilde\nabla_j\gets\frac{\sum_{i=1}^B|V_{ti}|(L_{V_{ti}}(\*w_{ti})-\lambda_t)1_{c_{ti}=c_j}}{{\theta_t}_{[j]}\sum_{i=1}^B|V_{ti}|}~\forall~j$\\
		$\theta_{t+1}\gets\theta_t\odot\exp(-\eta_t\tilde\nabla)$\\
		$\theta_{t+1}\gets\theta_{t+1}/\|\theta_{t+1}\|_1$
		
	}
	\KwOut{model $\*w$, hyperparameter distribution $\theta$}
	\caption{\label{alg:fedex}
		\FedEx
	}
\end{algorithm}
\end{minipage}
\begin{minipage}{0.45\linewidth}
	\centering
	\includegraphics[width=0.49\linewidth]{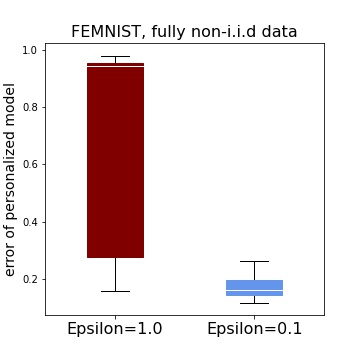}
	\hfill
	\includegraphics[width=0.49\linewidth]{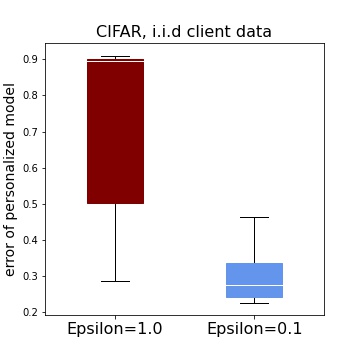}
	\caption{\label{fig:perturb}
		Comparison of the range of performance values attained using different perturbation settings.
		Although the range is much smaller for $\epsilon=0.1$ than for $\epsilon=1.0$ (the latter is the entire space), it still covers a large (roughly 10-20\%) range of different performance levels on both FEMNIST (left) and CIFAR (right).
	}
\end{minipage}
\end{figure}

We call this alternating method for solving \eqref{eq:stochastic} \FedEx and describe it for a general $\Alg_a$ consisting of sub-routines $\Agg_b$ and $\SGD_c$ in Algorithm~\ref{alg:fedex};
recall from Section~\ref{sec:federated} that many FL methods can be decomposed this way, so our approach is widely applicable.
\FedEx has a minimal overhead, consisting only of the last four lines of the outer loop updating $\theta$.
Thus, as with weight-sharing, \FedEx can be viewed as reducing the complexity of tuning local hyperparameters to that of training a single model.
Each update to $\theta$ requires a step-size $\eta_t$ and an approximation $\tilde\nabla$ of the gradient w.r.t. $\theta$;
for the latter we obtain an estimate $\tilde\nabla_j$ of each gradient entry via the reparameterization trick, whose variance we reduce by subtracting a baseline $\lambda_t$.
How we set $\eta_t$ and $\lambda_t$ is detailed in the Appendix.

To see how \FedEx is approximately optimizing the relaxed objective~\eqref{eq:stochastic}, we can consider the case where $\Alg_a$ is Reptile \citep{nichol:18}, which was designed to optimize some approximation of \eqref{eq:objective} for fixed $c$, or equivalently the relaxed objective for an atomic distribution $\D_\theta$.
The theoretical literature on meta-learning \citep{khodak:19a,khodak:19b} shows that Reptile can be interpreted as optimizing a surrogate objective minimizing the squared distance between $\*w$ and the optimum of each task $i$, with the latter being replaced by the last iterate in practice.
It is also shown that the surrogate objective is useful for personalization in the online convex setting.\footnote{Formally they study a sequence of upper bounds and not a surrogate objective, as their focus is online learning.}
As opposed to this past work, \FedEx makes two gradient updates in the outer loop, on two disjoint sets of variables: the first is the sub-routine $\Agg_b$ of $\Alg_a$ that aggregates the outputs of local training and is using the gradient of the surrogate objective, since the derivative of the squared distance is the difference between the initialization $\*w$ and the parameter at the last iterate of $\SGD_c$;
the second is the exponentiated gradient update that is directly using an unbiased estimate of the derivative of the second objective w.r.t. the distribution parameters $\theta$.
Thus, roughly speaking \FedEx runs simultaneous stochastic gradient descent on the relaxed objective \eqref{eq:stochastic}, although for the variables $\*w$ we are using a first-order surrogate.
In the theoretical portion of this work we employ this interpretation to show the approach works for tuning the step-size of online gradient descent in the online convex optimizations setting.

\vspace{-1mm}
\paragraph{Wrapping \FedEx}
We can view \FedEx as an algorithm of the form tuned by Algorithm~\ref{alg:baseline} that implements federated training of a supernet parameter $(\*w,\theta)$, with the local training routine $\SGD$ including a step for sampling $c\sim\D_\theta$ and the server aggregation routine including an exponentiated update of $\theta$.
Thus we can wrap \FedEx in Algorithm~\ref{alg:baseline}, which we find useful for a variety of reasons:
\begin{itemize}[leftmargin=*,topsep=-1pt,noitemsep]\setlength\itemsep{2pt}
	\item The wrapper can tune the settings of $b$ for the aggregation step $\Agg_b$, which \FedEx cannot.
	\item \FedEx itself has a few hyperparameters, e.g. how to set the baseline $\lambda_t$, which can be tuned.
	\item By running multiple seeds and potentially using early stopping, we can run \FedEx using more aggressive steps-sizes and the wrapper will discard cases where this leads to poor results.
	\item We can directly compare \FedEx to a regular hyperparameter optimization scheme run over the original algorithm, e.g. \FedAvg, by using the same scheme to both wrap \FedEx and tune \FedAvg.
	\item Using the wrapper allows us to determine the configurations $c_1,\dots,c_k$ given to Algorithm~\ref{alg:fedex} using a local perturbation scheme (detailed next) while still exploring the entire hyperparameter space.
\end{itemize}

\vspace{-1mm}
\paragraph{Local Perturbation}
It remains to specify how to select the configurations $c_1,\dots,c_k\in\Config$ to pass to Algorithm~\ref{alg:fedex}.
While the simplest approach is to draw from $\Unif^k(\Config)$, we find that this leads to unstable behavior if the configurations are too distinct from each other.
To interpolate between sampling $c_i$ independently and setting them to be identical (which would just be equivalent to the baseline algorithm), we use a simple {\em local perturbation} method in which $c_1$ is sampled from $\Unif(\Config)$ and $c_2,\dots,c_k$ are sampled uniformly from a local neighborhood of $\Config$.
For continuous hyperparameters (e.g. step-size, dropout) drawn from an interval $[a,b]\subset\R$ the local neighborhood is $[c\pm(b-a)\varepsilon]$ for some $\varepsilon\ge0$, i.e. a scaled $\varepsilon$-ball;
for discrete hyperparameters (e.g. batch-size, epochs) drawn from a set $\{a,\dots,b\}\subset\Z$, the local neighborhood is similarly $\{c-\lfloor(b-a)\varepsilon\rfloor,\dots,c+\lceil(b-a)\varepsilon\rceil\}$;
in our experiments we set $\varepsilon=0.1$, which works well, but run ablation studies varying these values in the appendix showing that a wide range of them leads to improvement.
Note that while local perturbation does limit the size of the search space explored by each instance of \FedEx, as shown in Figure~\ref{fig:perturb} the difference in performance between different configurations in the same ball is still substantial.

\vspace{-1mm}
\paragraph{Limitations of \FedEx}
While \FedEx is applicable to many important FL algorithms, those that cannot be decomposed into local fine-tuning and aggregation should instead be tuned by one of our baselines, e.g. SHA.
\FedEx is also limited in that it is forced to rely on such algorithms as wrappers for tuning its own hyperparameters and certain FL hyperparameters such as server learning rate.


\section{Theoretical Analysis for Tuning the Step-Size in an Online Setting}\label{sec:theory}
\vspace{-1mm}

As noted in Section~\ref{sec:ws}, \FedEx can be viewed as alternating minimization, with a gradient step on a surrogate personalization loss and an exponentiated gradient update of the configuration distribution $\theta$.
We make this formal and prove guarantees for a simple variant of \FedEx in the setting where the server has one client per round, to which the server sends an initialization to solve an online convex optimization (OCO) problem using online gradient descent (OGD) on a sequence of $m$ adversarial convex losses (i.e. one SGD epoch in the stochastic case).
Note we use ``client'' and ``task'' interchangeably, as the goal is a meta-learning (personalization) result.
The performance measure here is {\em task-averaged regret}, which takes the average over $\tau$ clients of the regret they incur on its loss:\vspace{-2mm}
\begin{equation}\label{eq:tar}
	\bar{\mathbf{R}}_\tau=\frac1\tau\sum_{t=1}^\tau\sum_{i=1}^m\ell_{t,i}(\*w_{t,i})-\ell_{t,i}(\*w_t^\ast)
\end{equation}
Here $\ell_{t,i}$ is the $i$th loss of client $t$, $\*w_{t,i}$ the parameter chosen on its $i$th round from a compact parameter space $\W$, and  $\*w_t^\ast\in\argmin_{\*w\in\W}\sum_{i=1}^m\ell_{t,i}(\*w)$ the task optimum.
In this setting, the Average Regret-Upper-Bound Analysis (ARUBA) framework \citep{khodak:19b} can be used to show guarantees for a \Reptile (i.e. \FedEx with a server step-size) variant in which at each round the initialization is updated as $\*w_{t+1}\gets(1-\alpha_t)\*w_t+\alpha_t\*w_t^\ast$ for server step-size $\alpha_t=1/t$.
Observe that the only difference between this update and \FedEx's is that the task-$t$ optimum $\*w_t^\ast$ is used rather than the last iterate of OGD on that task.
Specifically they bound task-averaged regret by\vspace{-0.5mm}
\begin{equation}\label{eq:aruba}
	\bar{\mathbf{R}}_\tau\le\tilde{\mathcal O}\left(\frac1{\sqrt[4]\tau}+V\right)\sqrt m\quad\textrm{for}\quad V^2=\min_{\*w\in\W}\frac1\tau\sum_{t=1}^\tau\|\*w-\*w_t^\ast\|_2^2
\end{equation}
Here $V$---the average deviation of the optimal actions $\*w_t^\ast$ across tasks---is a measure of {\em task-similarity}:
$V$ is small when the tasks (clients) have similar data and thus can be solved by similar parameters in $\W$ but  large when their data is different and so the optimum parameters to use are very different.
Thus the bound in \eqref{eq:aruba} shows that as the server (meta-learning) sees more and more clients (tasks), their regret on each decays with rate $1/\sqrt[4]{\tau}$ to depend only on the task-similarity, which is hopefully small if the client data is similar enough that transfer learning makes sense, in particular if $V\ll\diam(\W)$.
Since single-task regret has lower bound $\Omega(D\sqrt m)$, achieving asymptotic regret $V\sqrt m$ thus demonstrates successful learning of a useful initialization in $\W$ that can be used for personalization.
Note that such bounds can also be converted to obtain guarantees in the statistical meta-learning setting as well \citep{khodak:19b}.

A drawback of past results using the ARUBA framework is that they either assume the task-similarity $V$ is known in order to set the client step-size \citep{li:20c} or they employ an OCO method to learn the local step-size that cannot be applied to other potential algorithmic hyperparameters \citep{khodak:19b}.
In contrast, we prove results for using bandit exponentiated gradient to tune the client step-size, which is precisely the \FedEx update.
In particular, Theorem~\ref{thm:fedex} shows that by using a discretization of potential client step-sizes as the configurations in Algorithms~\ref{alg:fedex} we can obtain the following task-averaged regret:
\begin{Thm}\label{thm:fedex}
	Let $\W\subset\R^d$ be convex and compact with diameter $D=\diam(\W)$ and let $\ell_{t,i}$ be a sequence of $m\tau$ $b$-bounded convex losses---$m$ for each of $\tau$ tasks---with Lipschitz constant $\le G$.
	We assume that the adversary is oblivious within-task.
	Suppose we run Algorithm~\ref{alg:fedex} with $B=1$, configurations $c_j=\frac D{Gj\sqrt m}$ for each $j=1,\dots,k$ determining the local step-size of single-epoch SGD (OGD), $\*w_{ti}=\*w_t^\ast$, regret $\sum_{i=1}^m\ell_{t,i}(\*w_{t,i})-\ell_{t,i}(\*w_t)$ used in place of $L_{V_{ti}}(\*w_{ti})$, and $\lambda_t=0~\forall~t\in[\tau]$.
	Then if $\eta_t=\frac1{mb}\sqrt{\frac{\log k}{k\tau}}~\forall~t\in[\tau]$,  $k^\frac32=\frac{DG}b\sqrt{\frac\tau{2m}}$, and $\Agg_b(\*w,\*w_t^\ast)=(1-\alpha_t)\*w+\alpha_t\*w_t^\ast$ for $\alpha_t=1/t~\forall~t\in[\tau]$ we have (taking expectations over sampling from $D_{\theta_t}$)\vspace{-0.5mm}
	\begin{equation}\label{eq:thm}
	\E\bar{\mathbf{R}}_\tau\le\tilde{\mathcal O}\left(\sqrt[3]{m/\tau}+V\right)\sqrt m
	\end{equation}
\end{Thm}
The proof of this result, given in the supplement, follows the ARUBA framework of using meta OCO algorithm to optimize the initialization-dependent upper bound on the regret of OGD;
in addition we bound errors to the bandit setting and discretization of the step-sizes.
Theorem~\ref{thm:fedex} demonstrates that \FedEx is a sensible algorithm for tuning the step-size in the meta-learning setting where each task is an OCO problem, with the average regret across tasks (clients) converging to depend only on the task-similarity $V$, which we hope is small in the setting where personalization is useful.
As we can see by comparing to the bound in \eqref{eq:aruba}, besides holding for a more generally-applicable algorithm our bound also improves the dependence on $\tau$, albeit at the cost of an additional $m^\frac13$ factor.
Note that that the sublinear term can be replaced by $1/\sqrt\tau$ in the full-information setting, i.e. where required the client to try SGD with each configuration $c_j$ at each round to obtain regret for all of them.

\vspace{-1mm}
\vspace{20mm}
\section{Empirical Results}\label{sec:empirical}
\vspace{-1mm}

\setlength{\tabcolsep}{3pt}
\begin{table*}[!t]
	\centering
	\caption{\label{tab:results}\normalsize
		Final test error obtained when tuning using a standard hyperparameter tuning algorithm (SHA or RS) alone, or when using it for server (aggregation) hyperparameters while \FedEx tunes client (on-device training) hyperparameters. 
		The target model is the one used to compute on-device validation error by the wrapper method, as well as the one used to compute test error after tuning.
		Note that this table reports the final error results corresponding to the online evaluations reported in Figure~\ref{fig:online}, which measure performance as more of the computational budget is expended.
	}
	\vspace{0.5em}
	\begin{threeparttable}
		\footnotesize
		\scalebox{0.85}{
		\begin{tabular}{cccccccc}
			\toprule
			Wrapper & Target & Tuning & \multicolumn{2}{c}{Shakespeare} & \multicolumn{2}{c}{FEMNIST} & CIFAR-10 \\
			method & model &  method & i.i.d. & non-i.i.d. & i.i.d. & non-i.i.d. & i.i.d. \\
			\midrule
			& \multirow{2}{*}{global} & RS (server \& client) & $60.32 \pm 10.03$ & $64.36 \pm 14.19$ & $22.81 \pm 4.56$ & $22.98 \pm 3.41$ & $30.46 \pm 9.44  $\\
			Random && \quad+\quad\FedEx(client) & $53.94 \pm 9.13 $ & $57.70 \pm 17.57$ & $20.96 \pm 4.77$ & $22.30  \pm 3.66$ & $34.83 \pm 14.74$ \\
			\cline{2-8}
			Search & person- & RS (server \& client) & $61.10  \pm 9.32 $ & $61.71 \pm 9.08 $ & $17.45 \pm 2.82$ & $17.77 \pm 2.63$ & $34.89 \pm 10.56 $\\
			(RS) & alized & \quad+\quad\FedEx(client) & $ 54.90  \pm 9.97 $ & $56.48 \pm 13.60 $ & $16.31 \pm 3.77$ & $15.93 \pm 3.06$ & $39.13 \pm 15.13$ \\
			\midrule
			& \multirow{2}{*}{global} & SHA (server \& client) & $ 47.38 \pm 3.40  $ & $46.79 \pm 3.51 $ & $18.64 \pm 1.68$ & $20.30  \pm 1.66$ & $21.62 \pm 2.51 $ \\
			Successive && \quad+\quad\FedEx(client) & $\mathbf{44.52} \pm 1.68 $ & $\mathbf{45.24} \pm 3.31 $ & $19.22 \pm 2.05$ & $19.43 \pm 1.45$ & $\mathbf{20.82} \pm 1.37 $ \\
			\cline{2-8}
			Halving & person- & SHA (server \& client) & $46.77 \pm 3.61 $ & $48.04 \pm 3.72 $ & $\mathbf{14.79} \pm 1.55$ & $14.78 \pm 1.31$ & $24.81 \pm 6.13$  \\
			(SHA) & alized & \quad+\quad\FedEx(client) & $46.08 \pm 2.57 $ & $45.89 \pm 3.76 $ & $14.97 \pm 1.31$ & $\mathbf{14.76} \pm 1.70 $ & $21.77 \pm 2.83 $ \\
			\bottomrule
		\end{tabular}}
	\end{threeparttable}
\end{table*}

In our experiments, we instantiate \FedEx on the problem of tuning \FedAvg, \FedProx, and \Reptile; 
the first is the most popular algorithm for federated training, the second is an extension designed for heterogeneous devices, and the last is a compatible meta-learning method used for learning initializations for personalization.
At communication round $t$ these algorithms use the aggregation\vspace{-1mm}
\begin{equation}\label{eq:fedavg}
	\Agg_b(\*w,\{\*w_i\}_{i=1}^B)=(1-\alpha_t)\*w+\frac{\alpha_t}{\sum_{i=1}^B|T_{ti}|}\sum_{i=1}^B|T_{ti}|\*w_i
\end{equation}
for some learning rate $\alpha_t>0$ that can vary through time;
in the case of $\FedAvg$ we have $\alpha_t=1~\forall~t$.
The local training sub-routine $\SGD_c$ is SGD with hyperparameters $c$ over some objective defined by the training data $T_{ti}$, which can also depend on $c$. 
For example, to include \FedProx we include in $c$ an additional local hyperparameter for the proximal term compared with that of \FedAvg.
%

We tune several hyperparameters of both aggregation and local training;
for the former we tune the server learning rate schedule and momentum, found to be helpful for personalization \citep{jiang:19};
for the latter we tune the learning rate, momentum, weight-decay, the number of local epochs, the batch-size, dropout, and proximal regularization.
Please see the supplementary material for the exact hyperparameter space considered.
While we mainly evaluate \FedEx in cross-device federated settings, which is generally more difficult than cross-silo in terms of hyperparameter optimization, \FedEx can be naturally applied to cross-silo settings, where the challenges of heterogeneity, privacy requirements, and personalization remain.


Because our baseline is running Algorithm~\ref{alg:baseline}, a standard hyperparameter tuning algorithm, to tune all hyperparameters, and because we need to also wrap \FedEx in such an algorithm for the reasons described in Section~\ref{sec:ws}, our empirical results will test the following question:
does \FedEx, wrapped by random search (RS) or a successive halving algorithm (SHA), do better than RS or SHA run with the same settings directly?
Here ``better'' will mean both the final test accuracy obtained and the online evaluation setting, which tests how well hyperparameter optimization is doing at intermediate phases.
Furthermore, we also investigate whether \FedEx can improve upon the wrapper alone even when targeting a good {\em global} and not personalized model, i.e. when elimination decisions are made using the average global validation loss.
We run Algorithm~\ref{alg:baseline} on the personalized objective and use RS and SHA with elimination rate $\eta=3$, the latter following Hyperband \citep{li:18}.
To both wrappers we allocate the same (problem-dependent) tuning budget.
To obtain the elimination rounds  in Algorithm~\ref{alg:baseline} for SHA, we set the number of eliminations to $R=3$, fix a total communication round budget, and fix a maximum number of rounds to be allocated to any configuration $a$;
as detailed in the Appendix, this allows us to determine $T_1,\dots,T_R$ so as to use up as much of the budget as possible. 

\begin{figure*}[!t]
	\centering
	\includegraphics[width=0.329\linewidth]{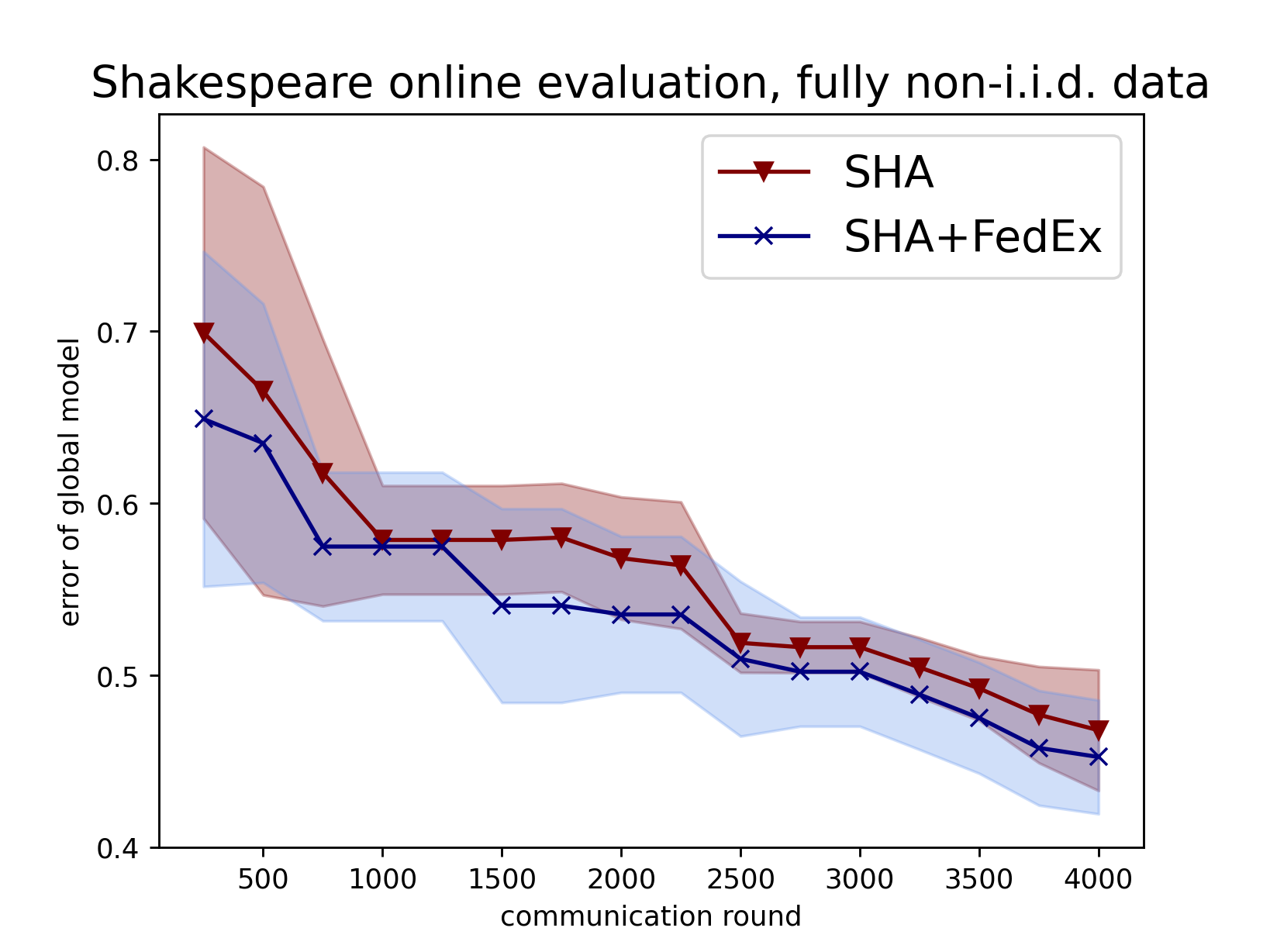}
	\includegraphics[width=0.329\linewidth]{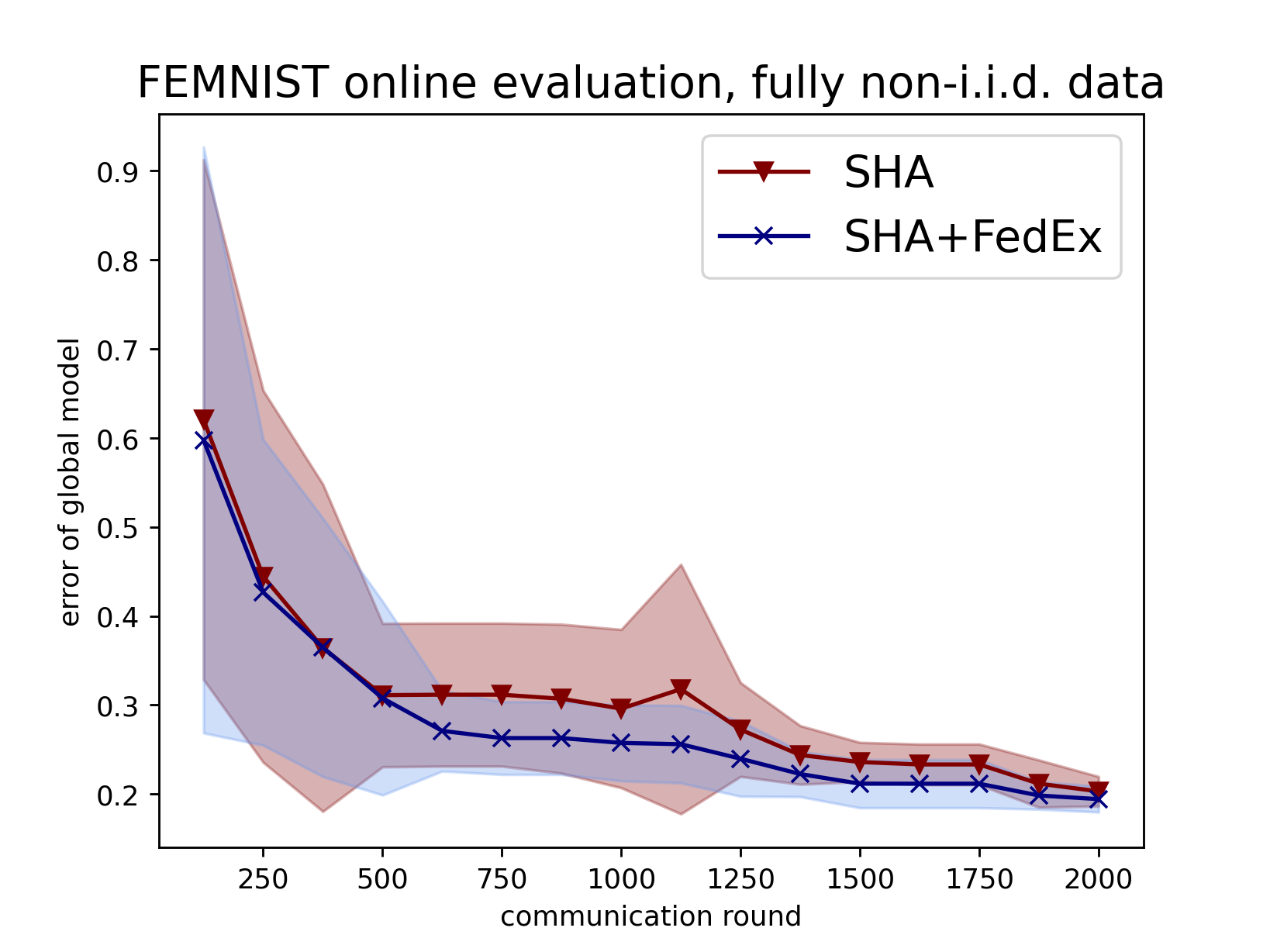}
	\includegraphics[width=0.329\linewidth]{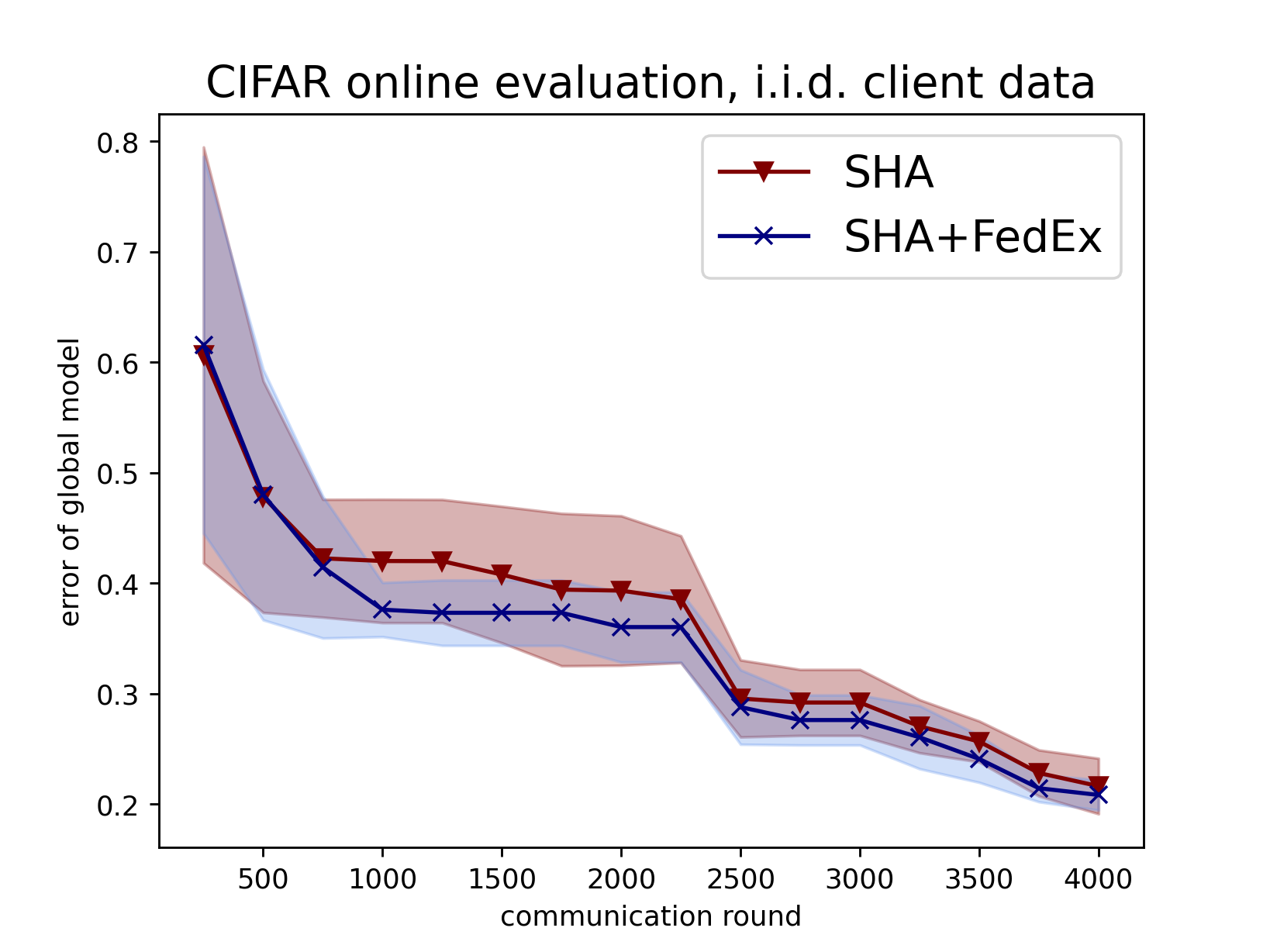}
	\includegraphics[width=0.329\linewidth]{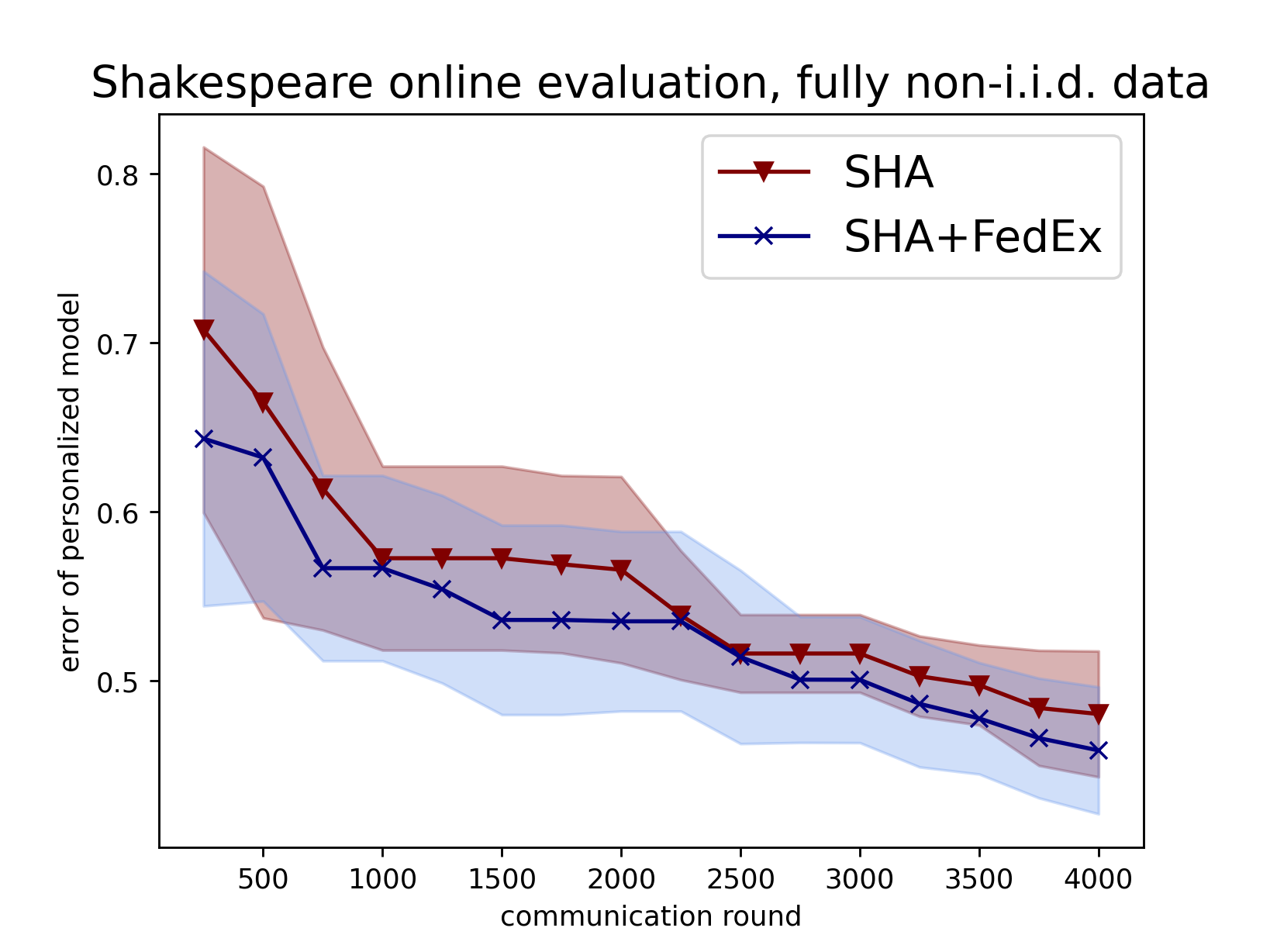}
	\includegraphics[width=0.329\linewidth]{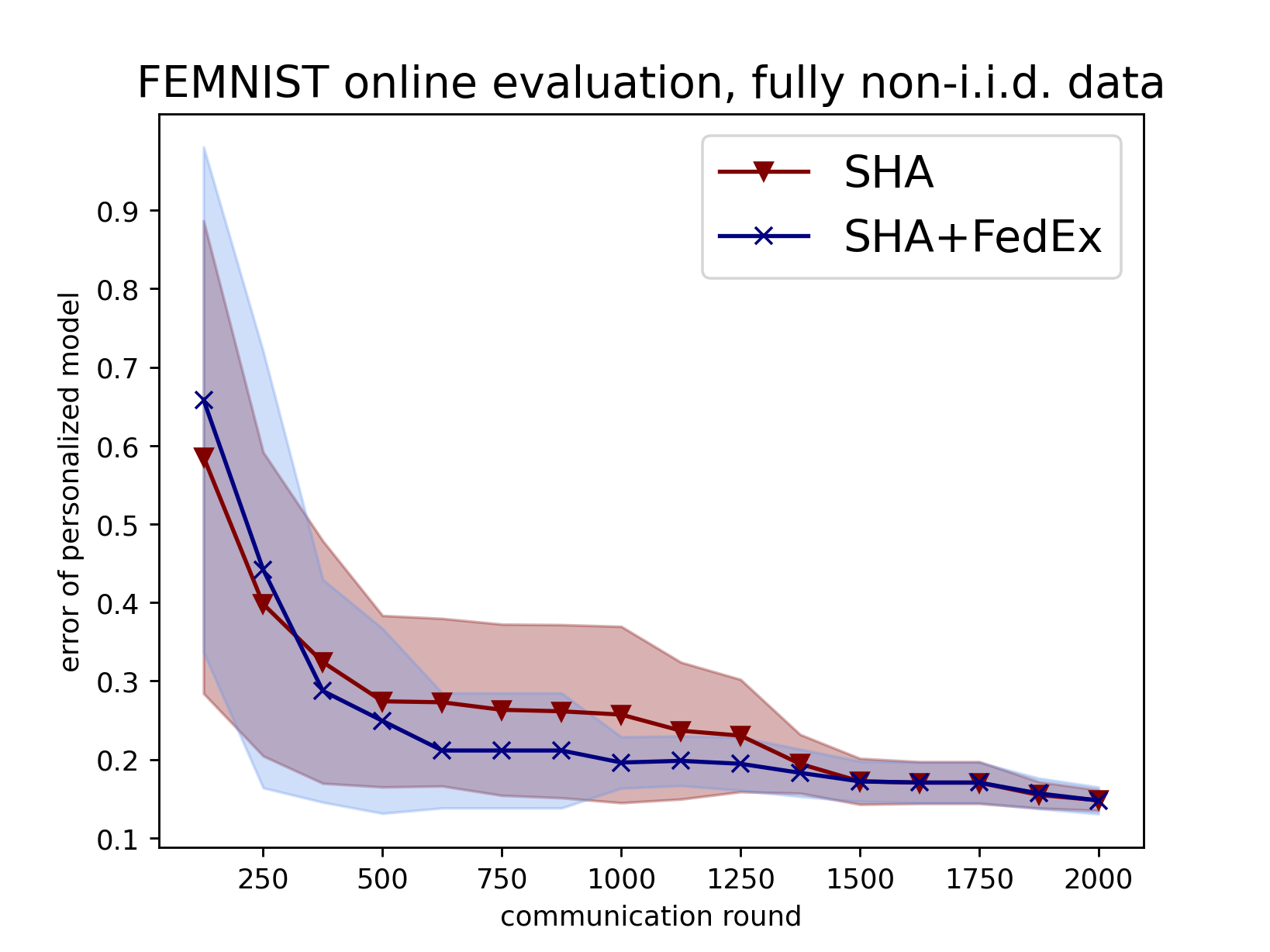}
	\includegraphics[width=0.329\linewidth]{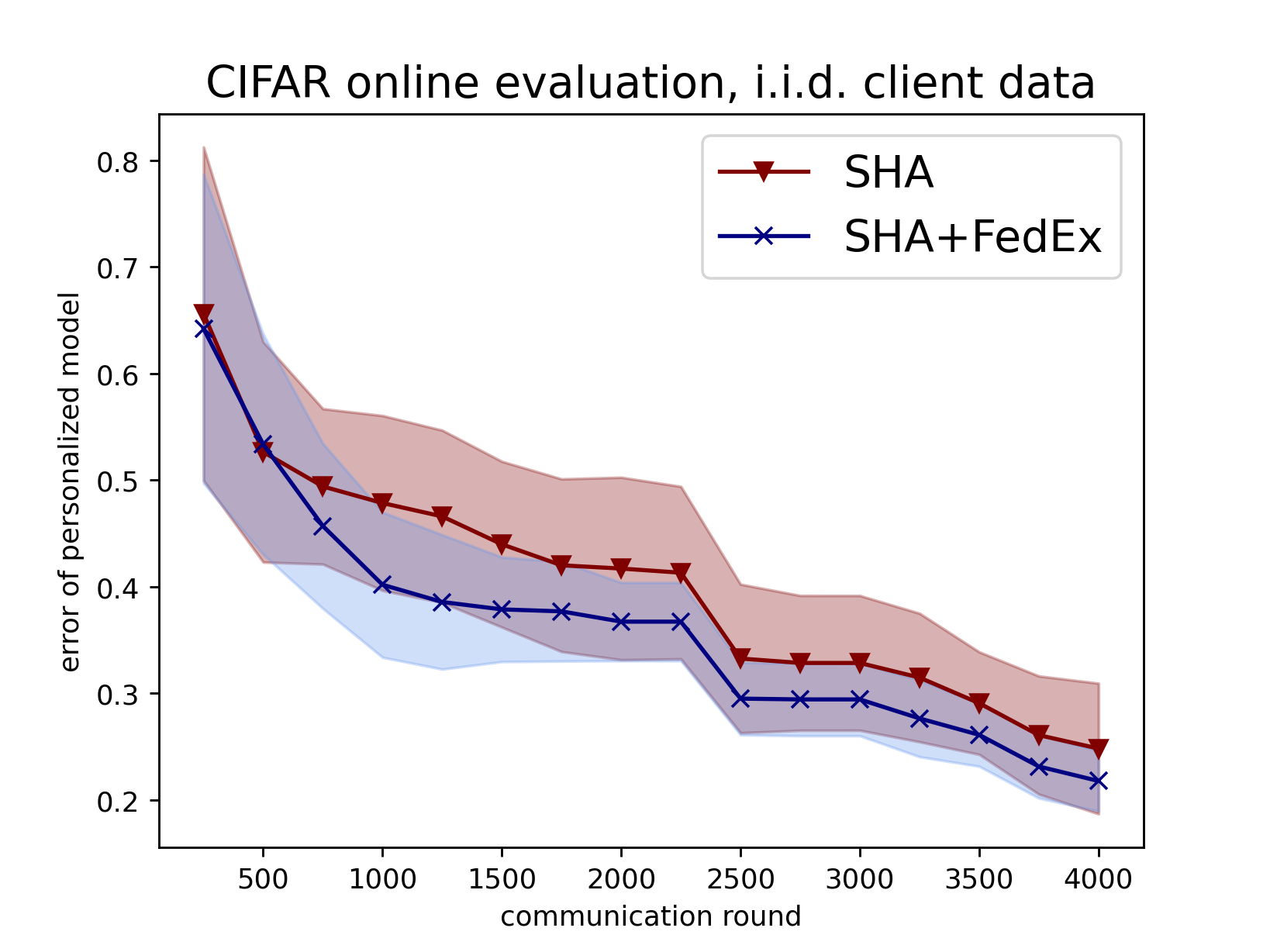}
	\caption{\label{fig:online}
		Online evaluation of \FedEx on the Shakespeare next-character prediction dataset (left), the FEMNIST image classification dataset (middle), and the CIFAR-10 image classification dataset (right) in the fully non-i.i.d. setting (except CIFAR-10).
		We report global model performance on the top and personalized performance on the bottom.
		All evaluations are run for three trials.
	}
\end{figure*}

We evaluate the performance of \FedEx on three datasets (Shakespeare, FEMNIST, and CIFAR-10) on both vision and language tasks.  We consider the following two different partitions of data: 
\begin{enumerate}[leftmargin=*,topsep=-1pt,noitemsep]\setlength\itemsep{2pt}
	\item Each device holds i.i.d. data. While overall data across the entire network can be non-i.i.d., we randomly shuffle local data \textit{within} each device before splitting into train, validation, and test sets.
	\item Each device holds non-i.i.d. data. In Shakespeare, each device is an actor and the local data is split according to the temporal position in the play;
	in FEMNIST, each device is the digit writer and the local data is split randomly;
	in CIFAR-10, we do not consider a non-i.i.d. setting. 
\end{enumerate}
For Shakespeare and FEMNIST we use 80\% of the data for training and 10\% each for validation and testing.
In CIFAR-10 we hold out 10K examples from the usual training/testing split for validation.
The backbone models used for Shakespeare and CIFAR-10 follow from the \FedAvg evaluation \citep{mcmahan:17} and use 4K communications rounds (at most 800 round for each arm), while that of FEMNIST follows from LEAF \citep{caldas:18} and uses 2K communication rounds (at most 200 for each arm).

Table~\ref{tab:results} presents our main results, displaying the final test error of the target model after tuning using either a wrapper algorithm alone or its combination with \FedEx.
The evaluation shows that using \FedEx on the client parameters is either equally or more effective in most cases;
in particular, a \FedEx-modified method performs best everywhere except i.i.d. FEMNIST, where it is very close.
Furthermore, \FedEx frequently improves upon the wrapper algorithm by 2 or more percentage points.

We further present online evaluation results in Figure~\ref{fig:online}, where we display the test error of \FedEx wrapped with SHA compared to SHA alone as a function of communication rounds.
Here we see that for most of training \FedEx is either around the same or better then the alternative, except at the beginning;
the former is to be expected since the randomness of \FedEx leads to less certain updates at initialization.
Nevertheless \FedEx is usually better than the SHA baseline by the halfway point.

\vspace{-1mm}
\section{Conclusion}\label{sec:conclusion}
\vspace{-1mm}

In this paper we study the problem of hyperparameter optimization in FL, starting with identifying the key challenges and proposing reasonable baselines that adapts standard approaches to the federated setting.
We further make a novel connection to the weight-sharing paradigm from NAS---to our knowledge the first instance of this being used for regular (non-architectural) hyperparameters---and use it to introduce \FedEx.
This simple, low-overhead algorithm for accelerating the tuning of hyperparameters in federated learning can be theoretically shown to successfully tune the step-size for multi-task OCO problems and effectively tunes \FedAvg, \FedProx, and \Reptile on standard benchmarks.
The scope of application of \FedEx is very broad, including tuning actual architectural hyperparameters rather than just settings of local SGD, i.e. doing federated NAS, and tuning initialization-based meta-learning algorithms such as \Reptile and \MAML.
Lastly, any work on FL comes with privacy and fairness risks due its frequent use of sensitive data;
thus any application of our work must consider tools being developed by the community for mitigating such issues~\citep{mcmahan2018learning,li2020fair}.

\vspace{-1mm}
\section*{Acknowledgments}
\vspace{-1mm}

This material is based on work supported by the National Science Foundation under grants CCF-1535967, CCF-1910321, IIS-1618714, IIS-1901403, SES-1919453, IIS-1705121, IIS-1838017, IIS-2046613 and IIS-2112471; the Defense Advanced Research Projects Agency under cooperative agreements HR00112020003 and FA875017C0141; an AWS Machine Learning Research Award; an Amazon Research Award; a Bloomberg Research Grant; a Microsoft Research Faculty Fellowship; an Amazon Web Services Award; a Facebook Faculty Research Award; funding from Booz Allen Hamilton Inc.; a Block Center Grant; and a Two Sigma Fellowship Award. Any opinions, findings and conclusions or recommendations expressed in this material are those of the author(s) and do not necessarily reflect the views of any of these funding agencies.

\bibliography{references}
\bibliographystyle{plain}

\appendix

\section{Proof of Theorem~\ref{thm:fedex}}

\begin{proof}
	Let $\gamma_t\sim\D_{\theta_t}$ be the step-size chosen at time $t$.
	Then we have that
	\begin{align*}
	\tau\E\bar{\mathbf{R}}_\tau
	&=\sum_{t=1}^\tau\E_{\gamma_t}\sum_{i=1}^m\ell_{t,i}\left(\*w_{t,i}^{(\*w_t,\gamma_t)}\right)-\sum_{i=1}^m\ell_{t,i}(\*w_t^\ast)\\
	&=\sum_{t=1}^\tau\sum_{j=1}^k{\theta_t}_{[j]}\sum_{i=1}^m\ell_{t,i}\left(\*w_{t,i}^{(\*w_t,c_j)}\right)-\sum_{i=1}^m\ell_{t,i}(\*w_t^\ast)\\
	&\le\frac{\log k}\eta+\eta k\tau m^2b^2\\
	&\quad+\min_{j\in[k]}\sum_{t=1}^\tau\sum_{i=1}^m\ell_{t,i}\left(\*w_{t,i}^{(\*w_t,c_j)}\right)-\min_{\*w\in\W}\sum_{i=1}^m\ell_{t,i}(\*w_t^\ast)\\
	&\le2mb\sqrt{\tau k\log k}+\min_{j\in[k]}\sum_{t=1}^\tau\frac1{2c_j}\|\*w_t-\*w_t^\ast\|_2^2+c_jmG^2\\
	&\le2mb\sqrt{\tau k\log k}+\min_{j\in[k]}\frac{D^2(1+\log \tau)}{2c_j}+\left(\frac{V^2}{2c_j}+c_jmG^2\right)\tau\\
	&\le2mb\sqrt{\tau k\log k}+\frac{D^2(1+\log \tau)+V^2\tau}{2\gamma^\ast}+\gamma^\ast mG^2\tau\\
	&\quad+\min_{j\in[k]}\left(\frac1{c_j}-\frac1{\gamma^\ast}\right)\frac{D^2(1+\log \tau)+V^2\tau}2+(c_j-\gamma^\ast)mG^2\tau\\
	&\le2mb\sqrt{\tau k\log k}+4D\sqrt{\frac{\tau+\tau\log \tau}2}+\left(2V+\frac Dk\right)G\tau\sqrt{\frac m2}\\
	&=mb\sqrt{2\tau\log\tau}+4D\sqrt{\frac{\tau+\tau\log \tau}2}+\left(DG+2GV\tau\right)\sqrt{\frac m2}
	\end{align*}
	where the second line uses linearity of expectations over $\gamma_t\sim\D_{\theta_t}$, the third substitutes the bandit regret of EG \citep[Corollary~4.2]{shalev-shwartz:11}, the fourth substitutes $\eta=\frac1{mb}\sqrt{\frac{\log k}{\tau k}}$ and the regret of OGD \citep[Corollary~2.7]{shalev-shwartz:11}, the fifth substitutes the regret guarantee of Adaptive OGD over functions $\frac12\|\*w_t-\*w_t^\ast\|_2^2$ \citep[Theorem~2.1]{bartlett:08} with step-size $\alpha_t=1/t$ and the definition of $V$, the sixth substitutes the best discretized step-size $c_j$ for the optimal $\gamma^\ast\in\left(0,\frac D{G\sqrt{2m}}\right]$, and the seventh substitutes $\frac V{2G\sqrt{2m}}+\frac D{2G}\sqrt{\frac{1+\log \tau}{2m\tau}}$ for $\gamma^\ast$ and $\argmin_{j:c_j\ge\gamma^\ast}$ for $\argmin_jc_j$.
	Setting $k^\frac32=\frac{DG}b\sqrt{\frac \tau{2m}}$ and dividing both sides by $\tau$ yields the result.
\end{proof}

\newpage
\section{Decomposing Federated Optimization Methods}

As detailed in Section~\ref{sec:federated} our analysis and use of \FedEx to tune local training hyperparameters depends on a formulation that decomposes FL methods into two subroutines:
a local training routine $\SGD_c(S,\*w)$ with hyperparameters $c$ over data $S$ and starting from initialization $\*w$ and an aggregation routine $\Agg_b$ with hyperparameters $b$.
In this section we discuss how a variety of federated optimization methods, including several of the best-known, can be decomposed in this manner.
This enables the application of \FedEx to tune their hyperparameters.

\subsection{\FedAvg \citep{mcmahan:17}}

The best-known FL method, \FedAvg runs SGD on each client in a batch starting from a shared initialization and then updates to the average of the last iterate of the clients, often weighted by the number of data points each client has.
The decomposition here is:
\begin{itemize}
	\item[$\SGD_c$] Local SGD (or another gradient-based algorithm, e.g. Adam \citep{kingma:15}), with $c$ being the standard hyperparameters such as step-size, momentum, weight-decay, etc.
	\item[$\Agg_b$] Weighted averaging, with no hyperparameters in $b$.
\end{itemize}

\subsection{\FedProx \citep{li:20b}}

\FedProx has the same decomposition as \FedAvg except local SGD is replaced by a proximal version that regularizes the routine to be closer to the initialization, adding another hyperparameter to $c$ governing the strength of this regularization.

\subsection{\Reptile \citep{nichol:18}}

A well-known meta-learning algorithm, \Reptile has the same decomposition as \FedAvg except the averaged aggregation is replaced by a convex combination of the initialization and the average of the last iterates, as in Equation~\ref{eq:fedavg}.
This adds another hyperparameter to $b$ governing the tradeoff between the two.

\subsection{\SCAFFOLD \citep{karimireddy:19}}

\SCAFFOLD comes in two variants, both of which compute and aggregate control variates in parallel to the model weights.
The decomposition here is:
\begin{itemize}
	\item[$\SGD_c$] Local SGD starting from a weight initialization with a control variate, which can be merged to form the local training initialization.
	The hyperparameters in $c$ are the same as in \FedAvg.
	\item[$\Agg_b$] Weighted averaging of both the initialization and the control variates, with the same hyperparameters as \Reptile.
\end{itemize}

\subsection{\texttt{FedDyn} \citep{acar2021federated}}

In addition to a \FedAvg/\FedProx/\Reptile-like training routine, this algorithm maintains a regularizer on each device that affects the local training routine.
While this statefulness cannot strictly be subsumed in our decomposition, since it does not introduce any additional hyperparameters the remaining hyperparameters can be tuned in the same manner as we do for \FedAvg/\FedProx/\Reptile.
In order to choose between using \texttt{FedDyn} or not, one can introduce a binary hyperparameter to $c$ specifying whether or not $\SGD_c$ uses that term in the objective it optimizes or not, allowing it also to be tuned via \FedEx.

\subsection{\texttt{FedPA} \citep{al-shedivat:21}}

This algorithm replaces local SGD in \FedAvg by a local Markov-chain Monte Carlo (MCMC) routine starting from the initialization given by aggregating the previous round's MCMC routines.
The decomposition is then just a replacement of local SGD and its associated hyperparameters by local MCMC and its hyperparameters, with the aggregation routine remaining the same.

\subsection{\texttt{Ditto} \citep{li2020ditto}}

Although it depends on what solver is used for the local solver and aggregation routines, in the simplest formulation, the local optimization of personalized models involves an additional regularization hyperparameter.
While the updating rule of \texttt{Ditto} is different from that of \FedProx, the hyperparameters can be decomposed and tuned in a similar manner.

\subsection{\MAML \citep{finn:17}}

A well-known meta-learning algorithm, \MAML takes one or more full-batch gradient descent (GD) steps locally and updates the global model using a second-order gradient using validation data.
The decomposition here is :
\begin{itemize}
	\item[$\SGD_c$] Local SGD starting from a weight initialization.
	The hyperparameters in $c$ are the same as in \FedAvg.
	The algorithm also returns second-order information required to compute the meta-gradient.
	\item[$\Agg_b$] Meta-gradient computation, summation, and updating using a standard descent method like Adam \citep{kingma:15}.
	The hyperparameters in $b$ are the hyperparameters of the latter.
\end{itemize}

\section{\FedEx Details}

\subsection{Stochastic Gradient used by \FedEx}

Below is a simple calculation showing that the stochastic gradient used to update the categorical architecture distribution of \texttt{FedEx} is an unbiased approximation of the true gradient w.r.t. its parameters.

\begin{align*}
\nabla_{\theta_j}&\E_{c_{ij}|\theta}L_{V_{ti}}(\*w_i)\\
&=\nabla_{\theta_j}\E_{c_{ij}|\theta}(L_{V_{ti}}(\*w_i)-\lambda)\\
&=\E_{c_{ij}|\theta}\left((L_{V_{ti}}(\*w_i)-\lambda)\nabla_{\theta_j}\log\mathbb P_\theta(c_{ij})\right)\\
&=\E_{c_{ij}|\theta}\left((L_{V_{ti}}(\hat w_k)-\lambda)\nabla_{\theta_j}\log\prod_{i=1}^n\mathbb P_\theta(c_{ij}=c_j)\right)\\
&=\E_{c_{ij}|\theta}\left((L_{V_{ti}}(\*w_i)-\lambda)\sum_{i=1}^n\nabla_{\theta_j}\log\mathbb P_\theta(c_{ij}=c_j)\right)\\
&=\E_{c_{ij}|\theta}\left(\frac{(L_{V_{ti}}(\*w_i)-\lambda)1_{c_{ij}=c_j}}{\theta_j}\right)
\end{align*}

Note that this use of the reparameterization trick has some similarity with a recent RL approach to tune the local step-size and number of epochs~\cite{mostafa:19};
however, \FedEx can be rigorously formulated as an optimization over the personalization objective, has provable guarantees in a simple setting, uses a different configuration distribution that leads to our exponentiated update, and crucially for practical deployment does not depend on obtaining aggregate reward signal on each round.

\subsection{\FedEx wrapped with SHA}
\newcommand{\algrule}[1][.2pt]{\par\vskip.2\baselineskip\hrule height #1\par\vskip.5\baselineskip}
For completeness, we present the pseudo code of wrapping \FedEx with SHA in Algorithm~\ref{alg:fedex_sha} below.
\begin{algorithm}[H]
	\DontPrintSemicolon
	
		\KwIn{distribution $\D$ over hyperparameters $\A$, elimination rate $\eta\in\mathbb N$, elimination rounds $\tau_0=0,\tau_1,\dots,\tau_R$}
		sample set of $\eta^R$ hyperparameters $H\sim\D^{[\eta^R]}$\\
		initialize a model $\*w_a\in\R^d$ for each $a\in H$\\
		\For{elimination round $r\in[R]$}{
			\For{setting $a=(b,c)\in H$}{
					 $s_a, \*w_a, \theta_a \gets$	\FedEx($\*w_a, b, c, \theta_a,   \tau_{r+1}-\tau_r$)
			}
			$H\gets\{a\in H:s_a\le\frac1\eta\textrm{-quantile}(\{s_a:a\in H\})\}$
		}
		\KwOut{remaining $a\in H$ and associated model $\*w_a$}
	 \algrule[.5pt]
	\FedEx($\*w, b, \{c_1,\dots,c_k\}, \theta,  \tau \geq 1$):\\
	\algrule
	initialize $\theta_1\gets \theta$\\
	 initialize shared weights $\*w_1 \gets \*w$\\
  \For{comm. round $t=1,\dots,\tau$}{
	  	\For{client $i=1,\dots,B$}{
			send $\*w_t,\theta_t$ to client\\
			 sample $c_{ti}\sim\D_{\theta_t}$\\
			 $\*w_{ti}\gets\SGD_{c_{ti}}(T_{ti},\*w_t)$\\
			 send $\*w_{ti},c_{ti},L_{V_{ti}}(\*w_{ti})$ to server
		 }
	  $\*w_{t+1}\gets\Agg_b(\*w,\{\*w_{ti}\}_{i=1}^B)$\\
	 set step size $\eta_t$ and baseline $\lambda_t$\\
		 $\tilde\nabla_j\gets\frac{\sum_{i=1}^B|V_{ti}|(L_{V_{ti}}(\*w_{ti})-\lambda_t)1_{c_{ti}=c_j}}{{\theta_t}_{[j]}\sum_{i=1}^B|V_{ti}|}~\forall~j$\\
		 $\theta_{t+1}\gets\theta_t\odot\exp(-\eta_t\tilde\nabla)$\\
		 $\theta_{t+1}\gets\theta_{t+1}/\|\theta_{t+1}\|_1$\\
		$s \gets \sum_{i=1}^B |V_{ti}| L_{V_{ti}} / \sum_{i=1}^B |V_{ti}|$
	}
	\quad  \textbf{Return} $s$, model $\*w$, hyperparameter distribution $\theta$ 
	\caption{\label{alg:fedex_sha}
		\FedEx wrapped with SHA
	}
\end{algorithm}

%
%
%

\subsection{Hyperparameters of \FedEx}

We tune the computation of the baseline $\lambda_t$, which we set to
$$\lambda_t=\frac1{\sum_{s<t}\gamma^{t-s}}\sum_{s<t}\frac{\gamma^{t-s}}{\sum_{i=1}^B|V_{ti}|}\sum_{i=1}^BL_{V_{ti}}(\*w_i)$$
for discount factor $\gamma\in[0,1]$.
As discussed in Section~\ref{sec:ws}, the local perturbation factor is set to $\varepsilon=0.1$. 27 configurations are used in each arm for SHA and RS. 
The number of configuration used per arm of \FedEx (i.e. the dimensionality of $\theta$) is the same (27).

\newpage
\section{Experimental Details}

Code implementing \FedEx is available at \url{https://github.com/mkhodak/fedex}.
The code automatically downloads CIFAR-10 data, while Shakespeare and FEMNIST data is made available by the LEAF repository: \url{https://github.com/TalwalkarLab/leaf}.

\subsection{Settings of the Baseline/Wrapper Algorithm}

We use the same settings of Algorithm~\ref{alg:baseline} for both tuning \FedAvg and for wrapping \FedEx.
Given an elimination rate $\eta$, number of elimination rounds $R$, resource budget $B$, and maximum rounds per arm $M$, we assign $T_1,\dots,T_R$ s.t.
$$T_i-T_{i-1}=\frac{T-M}{\frac{\eta^{n+1}-1}{\eta-1}-n-1}$$
(recall $T_0=0$) and assign any remaining resources to maximize resource use.
All remaining details were noted in Section~\ref{sec:empirical}.

\subsection{Hyperparameters of \FedAvg/\FedProx/\Reptile}
Server hyperparameters (learning rate $\alpha_t=\gamma^t$):
\begin{align*}
\log_{10}{\textrm{lr}}:&\qquad\Unif[-1,1] \\
\textrm{momentum}:&\qquad\Unif[0,0.9]\\
\log_{10}(1-\gamma):&\qquad\Unif[-4,-2]
\end{align*}
Local training hyperparameters (note we only use 1 epoch for Shakespeare to conserve computation):
\begin{align*}
\log_{10}(\textrm{lr}):&\qquad\Unif[-4,0]\\
\textrm{momentum}:&\qquad\Unif[0.0,1.0]\\
\log_{10}(\textrm{weight-decay}):&\qquad\Unif[-5,-1]\\
\textrm{epoch}:&\qquad\Unif\{1,2,3,4,5\}\\
\log_2(\textrm{batch}):&\qquad\Unif\{3,4,5,6,7\}\\
\textrm{dropout}:&\qquad\Unif[0,0.5]
\end{align*}

\section{Confidence Intervals}

\begin{table*}[!h]
	\centering
	\caption{\label{tab:conf}\normalsize
		Final test error obtained when tuning using a standard hyperparameter tuning algorithm (SHA or RS) alone, or when using it for server (aggregation) hyperparameters while \FedEx tunes client (on-device training) hyperparameters. 
		The target model is the one used to compute on-device validation error by the wrapper method, as well as the one used to compute test error after tuning.
		The confidence intervals displayed are 90\% Student-t confidence intervals for the mean estimates from Table~\ref{tab:results}, with 5 independent trials for Shakespeare, 10 for FEMNIST, 10 for RS on CIFAR, and 6 for SHA on CIFAR.
	}
	\vspace{0.5em}
	\begin{threeparttable}
		\footnotesize
		\scalebox{0.85}{
			\begin{tabular}{cccccccc}
				\toprule[\heavyrulewidth]
				Wrapper & Target & Tuning & \multicolumn{2}{c}{Shakespeare} & \multicolumn{2}{c}{FEMNIST} & CIFAR-10 \\
				method & model &  method & i.i.d. & non-i.i.d. & i.i.d. & non-i.i.d. & i.i.d. \\
				\midrule
				& \multirow{2}{*}{global} & RS (server \& client) & $60.32 \pm 9.56  $ & $ 64.36 \pm 13.53 $ & $ 22.81 \pm 2.64  $ & $ 22.98 \pm 1.98  $ & $ 30.46 \pm 5.47 $ \\
				Random && \quad+\quad\FedEx(client) & $53.94 \pm 8.70  $ & $ 57.70  \pm 16.75 $ & $ 20.96 \pm 2.77   $ & $ 22.30  \pm 2.12   $ & $ 34.83 \pm 8.54 $ \\
				\cline{2-8}
				Search & person- & RS (server \& client) & $61.10  \pm 8.89 $ & $ 61.71 \pm 8.66 $ & $ 17.45 \pm 1.63  $ & $ 17.77 \pm 1.52  $ & $ 34.89 \pm 6.12 $ \\
				(RS) & alized & \quad+\quad\FedEx(client) & $54.90  \pm 9.50 $ & $ 56.48 \pm 12.97 $ & $ 16.31 \pm 2.19  $ & $ 15.93 \pm 1.77  $ & $ 39.13 \pm 8.77 $  \\
				\midrule
				& \multirow{2}{*}{global} & SHA (server \& client) & $47.38 \pm 3.24 $ & $ 46.79 \pm 3.35 $ & $ 18.64 \pm 0.97 $ & $ 20.30  \pm 0.96 $ & $ 21.62 \pm 1.45 $ \\
				Successive && \quad+\quad\FedEx(client) & $\mathbf{44.52} \pm 1.60 $ & $ \mathbf{45.24} \pm 3.16 $ & $ 19.22 \pm 1.19  $ & $ 19.43 \pm 0.84 $ & $ \mathbf{20.82} \pm 0.79 $\\
				\cline{2-8}
				Halving & person- & SHA (server \& client) & $46.77 \pm 3.44 $ & $ 48.04 \pm 3.54 $ & $ \mathbf{14.79} \pm 0.90 $ & $ 14.78 \pm 0.75 $ & $ 24.81 \pm 3.55 $ \\
				(SHA) & alized & \quad+\quad\FedEx(client) & $46.08 \pm 2.45 $ & $ 45.89 \pm 3.58 $ & $ 14.97 \pm 0.76 $ & $ \mathbf{14.76} \pm 0.99$ & $ 21.77\pm 1.64 $ \\
				\bottomrule[\heavyrulewidth]
		\end{tabular}}
	\end{threeparttable}
\end{table*}

\newpage
\section{Ablation Studies}\label{sec:ablation}

%

\begin{figure}[!t]
\begin{minipage}{0.48\linewidth}
	\centering
	\includegraphics[width=\linewidth]{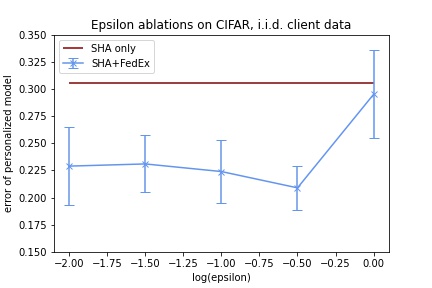}
	\caption{\label{fig:ablation}
		Comparison of different $\varepsilon$ settings for the local perturbation component of \FedEx from Section~\ref{sec:ws}.
	}
\end{minipage}
\hfill
\begin{minipage}{0.48\linewidth}
	\centering
	\includegraphics[width=\linewidth]{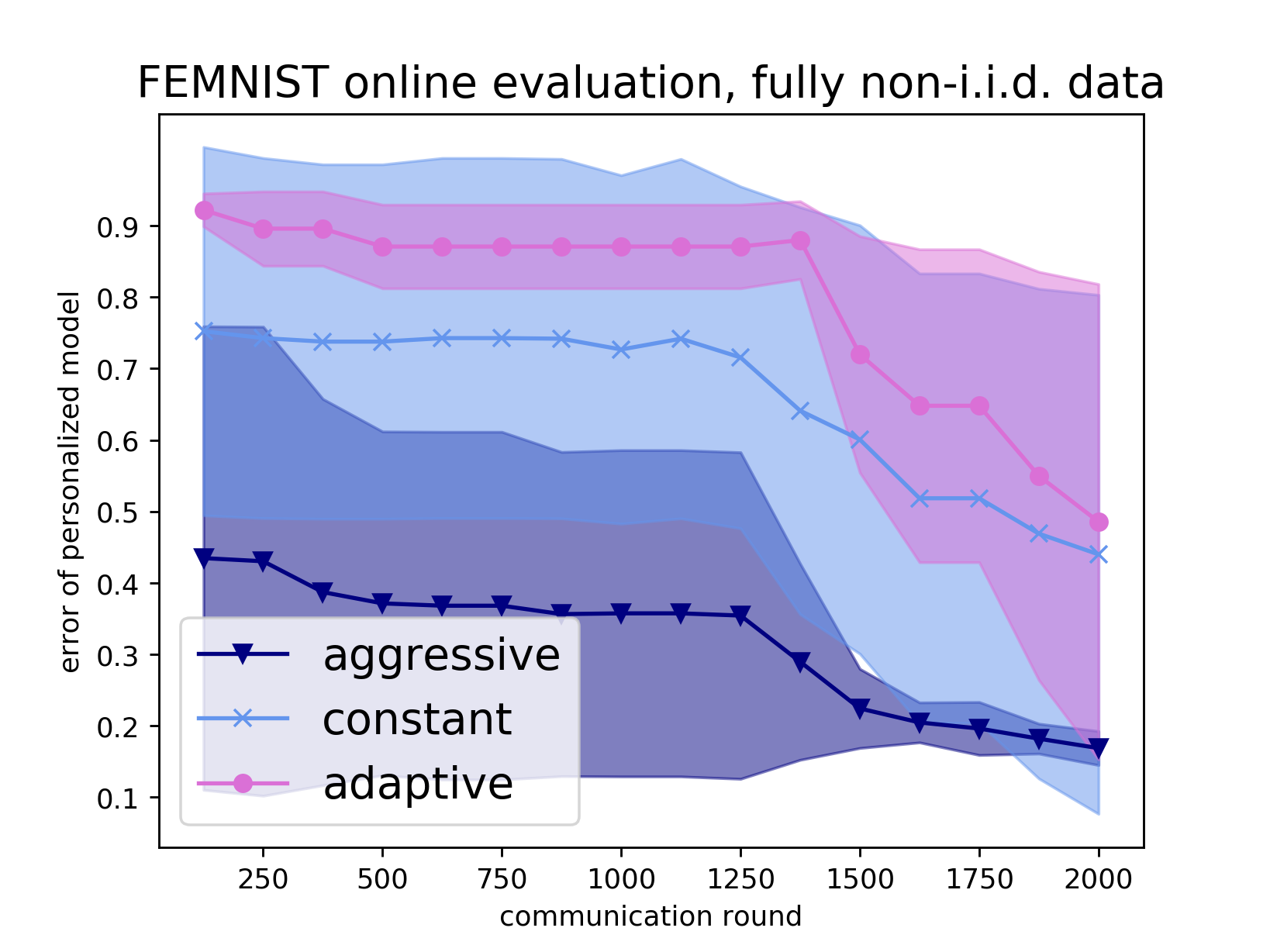}
	\caption{\label{fig:stepsize}
		Comparison of step-size schedules for $\eta_t$ in \FedEx.
		In practice we chose the `aggressive' schedule, which exhibits faster convergence to favorable configurations.
	}
\end{minipage}
\end{figure}

We now discuss two design choices of \FedEx and how they affect performance of the algorithm.
First, the choice of the local perturbation $\varepsilon=0.1$ discussed in Section~\ref{sec:ws};
we choose this setting due to its consistent performance across several settings.
In Figure~\ref{fig:ablation} we plot the performance of \FedEx on CIFAR-10 between $\varepsilon=0.0$ (no \FedEx, i.e. SHA only) and $\varepsilon=1.0$ (full \FedEx, i.e. client configurations are chosen independently) and show that while the use of a nonzero $\varepsilon$ is important, performance at fairly low values of $\varepsilon$ is roughly similar.

%

We further investigated the setting of the step-size $\eta_t$ for the exponentiated gradient update in \FedEx.
We examine three different approaches:
a constant rate of $\eta_t=\sqrt{2\log k}$, an `adaptive' schedule of $\eta_t=\sqrt{2\log k}/\sqrt{\sum_{s\le t}\|\tilde\nabla_s\|_\infty^2}$, and an `aggressive' schedule of $\eta_t=\sqrt{2\log k}/\|\tilde\nabla_t\|_\infty$.
Here $\tilde\nabla_t$ is the stochastic gradient w.r.t. $\theta$ computed in Algorithm~\ref{alg:fedex} at step $t$ and the form of the step-size is derived from standard settings for exponentiated gradient in online learning \cite{shalev-shwartz:11}.
We found that the `aggressive' schedule works best in practice, as shown in Figure~\ref{fig:stepsize}.
A key issue with using the `constant' and `adaptive' approaches is that they continue to assign high probability to several configurations late in the tuning process;
this slows down training of the shared weights.
One could consider a tradeoff between allowing \FedEx to run longer than while keeping the total budget constant, but for simplicity we chose the more effective `aggressive' schedule.

\end{document}